\pdfoutput=1
\let\arxiv\empty
\RequirePackage{fix-cm}
\documentclass[twocolumn,3p,a4paper,american,preprint,dvipsnames,x11names,svgnames]{elsarticle}
\usepackage[acronym]{glossaries}
\usepackage[detect-all,per-mode=symbol,range-units=single,binary-units]{siunitx}
\usepackage[inline]{enumitem}
\usepackage[T1]{fontenc}
\usepackage[utf8]{inputenc}
\usepackage{amsmath}
\usepackage{amssymb}
\usepackage{microtype}
\usepackage{babel}
\usepackage[babel]{csquotes}
\usepackage[l2tabu,orthodox]{nag}
\usepackage{booktabs}
\usepackage{mathtools}
\usepackage{bm}
\usepackage{tikz}
\usepackage{pgfplotstable}
\usepackage{pgfplots}
\usepackage{xfrac}
\usepackage{tabularx}
\usepackage{caption}
\usepackage{catchfile}
\usepackage{dblfloatfix}
\usepackage{natbib}
\usepackage{calc}
\usepackage{tikzscale}
\usepackage{tikzscalefix}
\usepackage{tikznonag}
\usepackage[labelformat=simple,subrefformat=simple]{subcaption}
\usepackage[all]{nowidow}
\usepackage{multirow}
\usepackage{pifont} 
\usepackage[abbreviations]{foreign}
\usepackage{rebuttal}
\usepackage{etoolbox}
\usepackage{sansmath}
\usepackage[switch,mathlines]{linenumbers}
\usepackage{varioref}
\usepackage{hyperref}
\usepackage[capitalise]{cleveref}

\newcommand{\cmark}{\textcolor{MediumSeaGreen}{\ding{51}}}
\newcommand{\xmark}{\textcolor{Firebrick3}{\ding{55}}}

\setlist*[enumerate]{label=(\arabic*)}

\newcolumntype{Y}{>{\centering\arraybackslash}X}

\graphicspath{{images/}}

\hypersetup{
  colorlinks,
  pdflang=en-US,
  pdfpagemode=UseNone,
  pdfpicktraybypdfsize=true,
  pdfprintscaling=None,
  pdfstartview=FitH,
  pdftitle={Automatic Classification of Defective Photovoltaic Module Cells in
       Electroluminescence Images},
  pdfauthor={Sergiu Deitsch},
  pdfkeywords={Deep learning, defect classification, electroluminescence
  imaging, photovoltaic modules, regression analysis, support vector machines,
  visual inspection},
}

\pgfplotsset{compat=1.13}

\usepgfplotslibrary{colorbrewer}
\usepgfplotslibrary{colormaps}
\usepgfplotslibrary{groupplots}
\usepgfplotslibrary{statistics}
\usepgfplotslibrary{units}

\usetikzlibrary{arrows.meta}
\usetikzlibrary{arrows}
\usetikzlibrary{backgrounds}
\usetikzlibrary{calc}
\usetikzlibrary{decorations.pathreplacing}
\usetikzlibrary{decorations}
\usetikzlibrary{external}
\usetikzlibrary{fit}
\usetikzlibrary{matrix}
\usetikzlibrary{pageref}
\usetikzlibrary{patterns}
\usetikzlibrary{positioning}

\tikzset{>=stealth'}

\glsdisablehyper

\newacronym{AGAST}{AGAST}{Adaptive and Generic Accelerated Segment Test}
\newacronym{AUC}{AUC}{Area Under the Curve}
\newacronym{BOW}{BOW}{Bag of Visual Words}
\newacronym{CAM}{CAM}{Class Activation Map}
\newacronym{CCD}{CCD}{Charge-coupled Device}
\newacronym{CNN}{CNN}{Convolutional Neural Network}
\newacronym{EL}{EL}{electroluminescence}
\newacronym{FAST}{FAST}{Features from Accelerated Segment Test}
\newacronym{FOV}{FOV}{Field-of-View}
\newacronym{GAP}{GAP}{Global Average Pooling}
\newacronym{GPU}{GPU}{Graphics Processing Unit}
\newacronym{HOG}{HOG}{Histogram of Oriented Gradients}
\newacronym{ICA}{ICA}{Independent Component Analysis}
\newacronym{IID}{IID}{independent and identically distributed}
\newacronym{IR}{IR}{infrared}
\newacronym{LED}{LED}{Light-emitting Device}
\newacronym{LO-RANSAC}{LO-RANSAC}{Locally Optimized Random Sample Consensus}
\newacronym{MCC}{MCC}{Matthews correlation coefficient}
\newacronym{MSE}{MSE}{Mean Squared Error}
\newacronym{PCA}{PCA}{Principal Component Analysis}
\newacronym{PHOW}{PHOW}{Pyramid Histogram of Visual Words}
\newacronym{PV}{PV}{photovoltaic}
\newacronym{RANSAC}{RANSAC}{Random Sample Consensus}
\newacronym{RBF}{RBF}{Radial Basis Function}
\newacronym{RMSE}{RMSE}{Root Mean Square Error}
\newacronym{ROC}{ROC}{Receiver Operating Characteristic}
\newacronym{SGD}{SGD}{Stochastic Gradient Descent}
\newacronym{SIFT}{SIFT}{Scale-invariant Feature Transform}
\newacronym{SSR}{SSR}{Signed Square Root}
\newacronym{SURF}{SURF}{Speeded Up Robust Features}
\newacronym{SVD}{SVD}{Singular Value Decomposition}
\newacronym{SVM}{SVM}{Support Vector Machine}
\newacronym{SVR}{SVR}{Support Vector Regression}
\newacronym{t-SNE}{\(t\)-SNE}{\(t\)-distributed Stochastic Neighbor Embedding}
\newacronym{VGG}{VGG}{Visual Geometry Group}
\newacronym{VLAD}{VLAD}{Vectors of Locally Aggregated Descriptors}

\newcommand\vggnet{\textsc{Vgg}-19}
\DeclareMathOperator*{\argmin}{argmin}
\DeclareMathOperator*{\sign}{sign}
\DeclarePairedDelimiter\abs{\lvert}{\rvert}

\let\vec\bm
\newcommand{\bmu}{\vec{\mu}}
\newcommand{\bx}{\vec{x}}

\newrobustcmd\rocbaselinemarker{%
  \tikzsetnextfilename{marker-roc-baseline}%
  \tikz [/pgfplots/every crossref picture] {\pgfplotsset{line legend,legend image code=roc baseline}}%
}
\newrobustcmd\outliermarker{%
  \tikzsetnextfilename{boxplot-outlier}%
  \tikz [/pgfplots/every crossref picture] {\pgfplotsset{outlier legend,legend image code={outlier mark,draw=none}}}%
}

\newcommand{\doi}[1]{%
  \begingroup
    \let\bibinfo\@secondoftwo
    \textsc{doi}:\,%
    \discretionary{}{}{}%
    \href{http://dx.doi.org/#1}{%
      \nolinkurl{#1}%
    }%
  \endgroup
}

\def\formatlabel#1{%
  \SI[round-mode=places,round-precision=2,zero-decimal-to-integer]{#1[1,2]}{\percent}%
  \\
  \( ( #1[1,1]\times#1[1,1] ) \)%
}

\def\placelabel[#1] #2:#3{%
  \node[pin={[#1]#2:\formatlabel#3}] at (axis cs:#3[1,1],#3[1,2]) {};
}

\pgfmathsetlengthmacro{\legendimageysep}{1pt}
\pgfmathsetlengthmacro{\crossrefyshift}{.5ex}

\pgfplotsset{
  place in a node/.code={
    \node[fit=(current bounding box), inner xsep=0pt,
          inner ysep=\legendimageysep, outer sep=0pt] {};
  },
  every crossref picture/.style={
    baseline,
    yshift=\crossrefyshift,
  },
}

\pgfplotsset{
  every axis legend/.append style={
    trim left=(current bounding box.west),
    trim right=(current bounding box.east)
  },
  every legend to name picture/.append style={
    trim left=(current bounding box.west),
    trim right=(current bounding box.east)
  },
}

\makeatletter

\pgfplotsset{
    groupplot xlabel/.initial={},
    every groupplot x label/.style={
        at={($({\pgfplots@group@name\space c1r\pgfplots@group@rows.west}|-{\pgfplots@group@name\space c1r\pgfplots@group@rows.outer south})!0.5!({\pgfplots@group@name\space c\pgfplots@group@columns r\pgfplots@group@rows.east}|-{\pgfplots@group@name\space c\pgfplots@group@columns r\pgfplots@group@rows.outer south})$)},
    },
    groupplot ylabel/.initial={},
    every groupplot y label/.style={
            rotate=90,
        at={($({\pgfplots@group@name\space c1r1.north}-|{\pgfplots@group@name\space c1r1.outer
west})!0.5!({\pgfplots@group@name\space c1r\pgfplots@group@rows.south}-|{\pgfplots@group@name\space c1r\pgfplots@group@rows.outer west})$)},
    },
    execute at end groupplot/.code={%
      \node [/pgfplots/every groupplot x label]
{\pgfkeysvalueof{/pgfplots/groupplot xlabel}};
      \node [/pgfplots/every groupplot y label]
{\pgfkeysvalueof{/pgfplots/groupplot ylabel}};
    }
}

\def\endpgfplots@environment@groupplot{%
    \endpgfplots@environment@opt%
    \pgfkeys{/pgfplots/execute at end groupplot}%
    \endgroup%
}

\@ifpackageloaded{hyperref}{%
  \pgfkeys{/tikz/disable hyperlinks/.code={%
    \let\ref\@refstar%
  }}}{%
  \pgfkeys{/tikz/disable hyperlinks/.code={}}%
}

\makeatother

\pgfkeys{/tikz/disable hyperlinks/.value forbidden}

\tikzset{caption/.code={%
  \tikzset{%
    outer sep=0pt,inner sep=0pt,
    anchor=north,align=left,text width=#1,
  }},
  caption/.default=\columnwidth,
}

\tikzset{legend column sep/.style={
    every even column/.append style={column sep=#1},
  },
  legend column sep/.default=1em,
}

\pgfplotsset{every axis legend/.append style={
  legend column sep
}}

\tikzset{defect score node/.code={
  \pgfmathsetmacro\usecolor{#1}%
  \pgfplotscolormapdefinemappedcolor\usecolor%
  \definecolor{mapped node color}{rgb}{\pgfmathresult}%
  \ifdim\usecolor pt > 500pt 
    \tikzset{color of colormap=0}%
    \colorlet{node text color}{.}%
  \else
    \tikzset{color of colormap=850}%
    \colorlet{node text color}{.}%
  \fi
  \tikzset{color of colormap=150}%
  \colorlet{outer}{.}%
  \tikzset{%
    node text color,
    fill=mapped node color,
    draw=outer,
    rounded corners=1.5,
    inner sep=1.5pt,
  }%
}}

\pgfplotsset{annotated module/.style={
  colormap/Reds,
  cycle list={[of colormap]},
  enlargelimits=false,
  axis equal image,
  y dir=reverse,
  hide axis,
  colorbar,
  point meta min=0,
  point meta max=100,
  colorbar style={
    ylabel near ticks,
    yticklabel pos=right,
    ylabel={Defect probability~[\si{\percent}]},
    ytick={0,25,...,100},
  },
  every node near coord/.append style={
    defect score node=\pgfplotspointmetatransformed,
    font=\tiny,
    opacity=0.65,
  },
  table/x=x,
  table/y=y,
  table/col sep=comma,
}}

\pgfplotsset{defect predictions/.code={
  \pgfkeys{/pgfplots/.cd,
    scatter,
    mark=none,
    draw=none,
    nodes near coords*={%
      \pgfmathfloattofixed{\pgfplotspointmeta}%
      \SI[round-mode=places,round-precision=2,zero-decimal-to-integer]{\pgfmathresult}{\percent}%
    },
    nodes near coords align={center},
    nodes near coords style={fill},
    point meta={100*\thisrow{probability}}
}}}

\pgfplotsset{4 by 4 confusion matrix/.style={
  width=5.25cm, height=5.25cm,
  enlargelimits=false,
  ylabel=Expected,
  xlabel=Predicted,
  ytick={0,...,3},
  yticklabel={\pgfkeys{/pgfplots/percent label format=\tick/3}},
  xtick={0,...,3},
  xticklabel={\pgfkeys{/pgfplots/percent label format=\tick/3}},
  x tick label style={rotate=45, anchor=north east},
  colormap/Blues-4,
  /tikz/adjust near node color/.code={%
    \pgfmathfloattofixed{\pgfplotspointmeta}%
    \ifdim\pgfmathresult pt > 50pt 
      \colorlet{node text color}{Blues-A}%
    \else
      \colorlet{node text color}{Blues-M}%
    \fi
    \tikzset{%
      node text color,
    }%
  },
  nodes near coords align={anchor=center},
  mesh/cols=4,
  mesh/rows=4,
  point meta=explicit,
  nodes near coords={%
    \pgfmathfloattofixed{\pgfplotspointmeta}%
    \SI[round-mode=places,round-precision=1]{\pgfmathresult}{\percent}%
  },
  every node near coord/.append style={adjust near node color},
  table/x=x,
  table/y=y,
  table/meta=count,
  table/col sep=comma,
}}

\pgfplotsset{2 by 2 confusion matrix/.style={
  width=5cm, height=5cm,
  enlargelimits=false,
  ylabel=Expected,
  xlabel=Predicted,
  ytick={0,1},
  yticklabel={\pgfkeys{/pgfplots/percent label format=\tick}},
  xtick={0,1},
  xticklabel={\pgfkeys{/pgfplots/percent label format=\tick}},
  y tick label style={rotate=90, anchor=south},
  x tick label style={rotate=0, anchor=north},
  /tikz/adjust near node color/.code={%
    \pgfmathfloattofixed{\pgfplotspointmeta}%
    \ifdim\pgfmathresult pt > 50pt 
      \tikzset{color of colormap=0}%
    \else
      \tikzset{color of colormap=850}%
    \fi
    \colorlet{node text color}{.}%
    \tikzset{%
      node text color,
    }%
  },
  nodes near coords align={anchor=center},
  mesh/cols=2,
  mesh/rows=2,
  point meta min=0,
  point meta max=100,
  point meta=explicit,
  nodes near coords={%
    \pgfmathfloattofixed{\pgfplotspointmeta}%
    \SI[round-mode=places,round-precision=1]{\pgfmathresult}{\percent}%
  },
  every node near coord/.append style={adjust near node color},
  table/x=x,
  table/y=y,
  table/meta=count,
  table/col sep=comma,
}}

\pgfplotsset{percent label format/.code={%
  \pgfmathparse{100*#1}%
  \SI[zero-decimal-to-integer,round-mode=places,round-precision=0]%
    {\pgfmathresult}{\percent}%
}}

\pgfplotsset{grid size format/.code={%
  \pgfkeys{/pgf/number format/.cd,int detect}%
  \pgfmathparse{#1}%
  \(\pgfmathprintnumber\pgfmathresult\times\pgfmathprintnumber\pgfmathresult\)%
}}

\pgfplotsset{confusion matrix/.style={%
  enlargelimits=false,
  xticklabel={\pgfkeys{/pgfplots/percent label format=\tick}},
  x tick label style={rotate=45, anchor=north east},
  yticklabel={\pgfkeys{/pgfplots/percent label format=\tick}},
  colormap/Blues-4,
}}

\pgfplotsset{confusion matrix base/.style={
}}

\pgfplotsset{confusion matrix group plot/.style={
  confusion matrix base,
  group/xlabels at=edge bottom,
  group/xticklabels at=edge bottom,
  group/ylabels at=edge left,
  group/yticklabels at=edge left,
}}

\pgfplotsset{wu/.style={
  legend pos=south east,
  width=10cm,
  height=7cm,
  xlabel=Sampling [cells],
  ylabel=\( F_1 \)~score~[\%],
  enlargelimits=0.05,
  x tick label style={rotate=45},
  xtick={5,15,...,75},
  minor x tick num=1,
  xticklabel=\pgfkeys{/pgfplots/grid size format=\tick},
}}

\pgfplotsset{dense conf group base/.style={
  wu,
  axis y discontinuity=crunch,
  ymin=55,
  ymax=77,
  width=\columnwidth+2em,height=(\columnwidth+2em)*3/4,
  table/col sep=comma,
  table/x expr=5*\lineno,
  group/xlabels at=edge bottom,
  group/ylabels at=edge left,
  group/vertical sep=6em,
  xlabel=,
  minor y tick num=1,
  legend columns=-1,
  /tikz/every pin/.append style={align=center,font=\scriptsize},
  groupplot xlabel={Grid size [\(n\times n\)~cells]},
}}

\pgfplotsset{dense conf group/.style={
  dense conf group base,
  title style={text width=\columnwidth},
  group/group size={2 by 2},
}}

\pgfplotsset{/pgfplots/table/PRF/.style={
  PRFbase,
  columns/features/.style={column name=Features, string type, column type=l, mathtext},
  columns/precision-micro/.style={column name=Precision, score},
  columns/recall-micro/.style={column name=Recall, score},
  columns/f1-micro/.style={column name=\( F_1 \)~score, score},
  columns/precision-macro/.style={column name=Precision, score},
  columns/recall-macro/.style={column name=Recall, score},
  columns/f1-macro/.style={column name=\( F_1 \)~score, score},
  columns={features, precision-micro, recall-micro, f1-micro, precision-macro, recall-macro, f1-macro},
  every head row/.style={
    before row={
      \toprule
      & \multicolumn{6}{c}{Micro average}
      & \multicolumn{6}{c}{Macro average}
      \\
    },
    after row=\midrule
  },
  every last row/.style={after row=\bottomrule}
}}

\pgfplotsset{/pgfplots/table/PRFbase/.style={
  score/.style={
    dec sep align,
    fixed,
    precision=2,
    fixed zerofill=true,
  },
  mathtext/.style={
    string replace*={5x5}{\( 5 \times 5 \)},
    string replace*={10x10}{\( 10 \times 10 \)},
    string replace*={15x15}{\( 15 \times 15 \)},
    string replace*={20x20}{\( 20 \times 20 \)},
    string replace*={25x25}{\( 25 \times 25 \)},
    string replace*={30x30}{\( 30 \times 30 \)},
    string replace*={35x35}{\( 35 \times 35 \)},
    string replace*={40x40}{\( 40 \times 40 \)},
    string replace*={45x45}{\( 45 \times 45 \)},
    string replace*={50x50}{\( 50 \times 50 \)},
    string replace*={55x55}{\( 55 \times 55 \)},
    string replace*={60x60}{\( 60 \times 60 \)},
    string replace*={65x65}{\( 65 \times 65 \)},
    string replace*={70x70}{\( 70 \times 70 \)},
    string replace*={75x75}{\( 75 \times 75 \)},
  },
  col sep=comma,
  text special chars=_,
}}

\pgfplotsset{/pgfplots/table/PRFuw/.style={
  PRFbase,
  columns/features/.style={column name=Sampling, string type, column type=l, mathtext},
  columns/precision-macro-unweighted/.style={column name=Precision, score},
  columns/recall-macro-unweighted/.style={column name=Recall, score},
  columns/f1-macro-unweighted/.style={column name=\( F_1 \)~score, score},
  columns/precision-macro-weighted/.style={column name=Precision, score},
  columns/recall-macro-weighted/.style={column name=Recall, score},
  columns/f1-macro-weighted/.style={column name=\( F_1 \)~score, score},
  columns={%
    features,
    precision-macro-unweighted, recall-macro-unweighted, f1-macro-unweighted,
    precision-macro-weighted, recall-macro-weighted, f1-macro-weighted
  },
  every head row/.style={
    before row={
      \toprule
      & \multicolumn{6}{c}{Unweighted}
      & \multicolumn{6}{c}{Weighted}
      \\
    },
    after row=\midrule
  },
  every last row/.style={after row=\bottomrule}
}}

\newcommand\AUC[2][]{%
  {#1\num[round-precision=2]{#2}}%
}

\pgfkeys{/auc/legend/title/.initial={AUC~[\%]:}}

\tikzset{roc baseline/.style={
  darkgray,
  opacity=0.75,
  thin,
  densely dashed,
}}

\pgfplotsset{plot roc baseline/.style={
  every axis/.append style={
    execute at begin axis={
      \draw[samples=100,smooth,domain=0.1:100,roc baseline] plot (\x, \x);
    },
  }
}}

\pgfplotsset{roc plot/.style={
  plot roc baseline,
  xmin=0.25,
  xmax=100,
  ymin=0,
  ymax=100,
  enlargelimits=0.05,
  xlabel=False positive rate~[\%],
  ylabel=True positive rate~[\%],
  cycle list name=color list,
  unbounded coords=jump,
  every axis plot/.append style={very thick},
  legend pos=outer north east,
  legend style={cells={anchor=west}},
  log basis x=10,
  x tick label style={%
    log ticks with fixed point,
  },
  log origin=0,
  table/col sep=comma,
  table/x=fpr,
  table/y=tpr,
  xmajorgrids=true,
  title style={text width=\columnwidth},
}}

\pgfplotsset{group roc plot/.style={
  roc plot,
  /pgfplots/xmode=log,
  group/group size=2 by 2,
  group/xlabels at=edge bottom,
  group/ylabels at=edge left,
  group/vertical sep=4em,
  xlabel=,
  groupplot xlabel=False positive rate~[\%],
  legend columns=4,
  legend style={font=\footnotesize},
}}

\tikzset{inner legend/.style={
  anchor=north west,font=\scriptsize,align=left,
  outer sep=1.25ex,
  disable hyperlinks,
  fill=white,fill opacity=0.75,draw=gray!50,
  text opacity=1,
  execute at begin node={%
    \tabcolsep=.5ex
  },
}}

\tikzset{plot caption/.code args={on #1 c#2r#3}{
  \tikzset{at={(#1 c#2r#3.south |- #1 c#2r#3.outer south)}}
}}

\tikzset{best dense/.style={font=\scriptsize\bfseries}}

\pgfplotsset{group roc mono poly/.style={
    scale only axis,
    width=.6\columnwidth,
    height=.5\columnwidth,
    roc plot,
    xmode=log,
    axis y discontinuity=none,
    ymin=0,
    ytick={0,25,50,75,100},
    legend pos=outer north east,
    group style={
      group size={3 by 1},
      horizontal sep=1em,
    },
    table/search path={data},
    group/xlabels at=edge bottom,
    group/xticklabels at=edge bottom,
    group/ylabels at=edge left,
    group/yticklabels at=edge left,
    minor y tick num=1,
    xlabel=,
    legend columns=3,
    groupplot xlabel={False positive rate~[\%]},
    title style={text width=.5\columnwidth},
    plot roc baseline,
}}

\makeatletter

\tikzset{canvas picture/.style={path picture={%
  \noexpand\pgfplotsextra{%
    \pgfmathsetlengthmacro{\picwidth}{%
      (\pgfkeysvalueof{/pgfplots/xmax}-\pgfkeysvalueof{/pgfplots/xmin})}%
    \pgfmathsetlengthmacro{\picheight}{%
      (\pgfkeysvalueof{/pgfplots/ymax}-\pgfkeysvalueof{/pgfplots/ymin})}%
    \pgfplotspointaxisdirectionxy{\picwidth}{\picheight}%
    \pgfgetlastxy{\xcoord}{\ycoord}%
    \pgfpathmoveto{\pgfplotspointaxisdirectionxy{0}{10}}%
    \ifdim\xcoord<\z@
      \pgfmathsetlengthmacro{\raisex}{\xcoord}%
      \pgfmathsetlengthmacro{\xcoord}{-\xcoord}%
    \else
      \pgfmathsetlengthmacro{\raisex}{\z@}%
    \fi
    \ifdim\ycoord<\z@
      \pgfmathsetlengthmacro{\raisey}{\ycoord}%
      \pgfmathsetlengthmacro{\ycoord}{-\ycoord}%
    \else
      \pgfmathsetlengthmacro{\raisey}{\z@}%
    \fi
    \kern\raisex\raisebox{\raisey}{\includegraphics[width=\xcoord,height=\ycoord]{#1}}%
  }%
}}}

\makeatother

\pgfplotsset{
  group boxplots/.style args={#1 by #2}{
    boxplot={
      draw position={1/(#2+1)+floor(\plotnumofactualtype/#1)+1/(#2+1)*mod(\plotnumofactualtype,#1)},
    },
  }
}

\colorlet{f1 score color}{GnBu-G!50!black}
\colorlet{auc score color}{OrRd-M}
\colorlet{accuracy color}{PuBu-M}

\pgfmathsetlengthmacro\outliermarksize{0.4ex}

\tikzset{
  f1 score colors/.style={pattern=crosshatch,draw=f1 score color,pattern color=GnBu-E!25!black},
  auc score colors/.style={pattern=north west lines,draw=auc score color,pattern color=OrRd-H!50},
  accuracy colors/.style={pattern=crosshatch dots,draw=accuracy color,pattern color=PuBu-H!50},
  outlier mark/.style={
    mark options={color=orange!35,draw=orange,solid},
    mark=otimes*,
    mark size=\outliermarksize,
  },
}

\pgfplotsset{outlier legend/.style={
  legend image code/.code={
    \node[inner sep=0pt,minimum width=2*\outliermarksize+.5pt,minimum height=2*\outliermarksize+.5pt]
      at (\outliermarksize,0pt) {};
    \draw [mark repeat=2,mark phase=2,##1]
      plot coordinates {
        (0cm,0cm)
        (\outliermarksize,0cm)
        (2*\outliermarksize,0cm)
    };
  }},
}

\pgfplotsset{
  f1 score/.style={
    boxplot/every box/.append style={f1 score colors},
    boxplot/every median/.append style={GnBu-E!50!black},
    outlier mark,
  },
  auc score/.style={
    boxplot/every box/.append style={auc score colors},
    boxplot/every median/.append style={OrRd-M},
    outlier mark,
  },
  accuracy/.style={
    boxplot/every box/.append style={accuracy colors},
    boxplot/every median/.append style={PuBu-M},
    outlier mark,
  },
}

\tikzset{
  scores baseline/.style={
    /pgfplots/.cd,
    no marks,
    cycle multiindex* list={
      dashed,densely dotted,densely dash dot\nextlist
      f1 score color,auc score color,accuracy color\nextlist
    },
    /tikz/.cd,
    semithick,
  },
}

\tikzset{
  subfigures/.code={
    \setcounter{subfigure}{0}
  }
}

\colorlet{baseline color}{Oranges-I}
\colorlet{CNN color}{Blues-I}

\pgfplotscreateplotcyclelist{baseline CNN comparison}{
  {baseline color},
  {CNN color}%
}

\tikzset{
  classifier group/.style={
    every node/.style={
      font=\bfseries,
      text depth=0.5ex,
    },
  },
  classifier group label/.style={
    decorate,decoration={brace,mirror},
  },
  eval/baseline/.style={/pgfplots/.cd, colormap/Oranges-4, /tikz/.cd},
  eval/CNN/.style={/pgfplots/.cd, colormap/Blues-4, /tikz/.cd},
  baseline group/.style={
    classifier group label,
    baseline color,
  },
  CNN group/.style={
    classifier group label,
    CNN color,
  },
}

\pgfplotsset{
  boxplot/draw/whisker/.code 2 args={
    \draw [/pgfplots/boxplot/every whisker/.try,thin]
      (boxplot cs:#1) -- (boxplot cs:#2);
    \draw [thick,/pgfplots/boxplot/every whisker/.try,solid]
      (boxplot whisker cs:#2,0) -- (boxplot whisker cs:#2,1);
  },
  boxplot/every box/.append style={black},
  boxplot/every median/.append style={very thick},
  boxplot/every whisker/.append style={gray},
  boxplot/every box/.append style={solid},
  boxplot/every median/.append style={solid},
  boxplot/every whisker/.append style={solid},
}

\pgfplotstableset{
  row sum/.style={
    create col/assign/.code={
      \def\entry{}
      \foreach \i in {#1} {
        \ifnum\pgfplotstablerow=\i
          \def\colsum{0}
          \pgfmathtruncatemacro\maxcolindex{\pgfplotstablecols-1}
          \pgfplotsforeachungrouped \col in {0,...,\maxcolindex}{
            \pgfmathsetmacro\colsum{\colsum+\thisrowno{\col}}
          }
          \xdef\entry{\colsum}
        \fi
      }
      \pgfkeyslet{/pgfplots/table/create col/next content}\entry
    }
  },
}

\pgfplotsset{colormap/Dark2-8}%
\pgfplotsset{colormap/Paired}%

\def\basecolorset{Set1}

\colorlet{gap outline}{\basecolorset-A}
\colorlet{gap fill}{gap outline!25}

\colorlet{fc outline}{\basecolorset-B}
\colorlet{fc fill}{fc outline!25}

\colorlet{activations outline}{\basecolorset-C}
\colorlet{activations fill}{activations outline!25}

\colorlet{conv outline}{gray}
\colorlet{conv fill}{gray!15}

\colorlet{max pooling outline}{\basecolorset-D}
\colorlet{max pooling fill}{max pooling outline!25}

\tikzset{custom/.style={
    font=\bfseries\boldmath,
}}

\tikzset{custom layer/.style={
    solid,
}}

\tikzset{block/.style={
    fill=white,
    line join=round,
    miter limit=1,
}}

\tikzset{conv block/.style={
    block,
    fill=conv fill,
    draw=conv outline,
}}

\tikzset{max pooling/.style={
    block,
    fill=max pooling fill,
    draw=max pooling outline,
}}

\tikzset{fc/.style={
    block,
    fill=fc fill,
    draw=fc outline,
}}

\tikzset{activations/.style={
    block,
    fill=activations fill,
    draw=activations outline,
}}

\tikzset{custom block label/.style={
    rotate=45,
    anchor=north east,
}}

\tikzset{block label/.style={
    custom block label,
    gray,
}}

\tikzset{block label line/.style={
    decorate,decoration={brace,mirror},
}}

\tikzset{gap/.style={
    draw=gap outline,
    fill=gap fill,
}}

\tikzset{gap block label/.style={
    custom block label,
    gap outline,
}}

\tikzset{fc1 block label/.style={
    block label,
    custom,
    fc outline,
}}

\tikzset{fc2 block label/.style={
    fc1 block label,
}}

\tikzset{activations block label/.style={
    block label,
    custom,
    activations outline,
}}

\tikzset{custom label line/.style={
    block label line,
    thick,
}}

\tikzset{gap label line/.style={
    custom label line,
    gap outline,
}}

\tikzset{fc1 label line/.style={
    custom label line,
    fc outline,
}}

\tikzset{fc2 label line/.style={
    custom label line,
    fc outline,
}}

\tikzset{activations label line/.style={
    custom label line,
    activations outline,
}}

\tikzset{input label/.style={
  align=center,
  rotate=90,
  anchor=south,
}}

\tikzset{input label line/.style={
  ->
}}

\tikzset{external/system call={pdflatex \tikzexternalcheckshellescape -halt-on-error -interaction=batchmode -jobname "\image" "\texsource"}}
\tikzset{external/mode=list and make}

\tikzsetexternalprefix{tikz/}
\begin{document}

\sisetup{
  round-mode=places,
  round-precision=2,
  add-integer-zero=false,
  zero-decimal-to-integer,
  group-minimum-digits=3,
  group-separator={{,}},
}


\pgfplotsset{every axis legend/.append style={
  legend column sep=1em,
}}

\ifx\arxiv\undefined\else
\rebuttalset{marked=false}
\fi

\begin{frontmatter}

\ifx\arxiv\undefined
\input{rebuttal}
\fi

\phantomsection
\pdfbookmark{Manuscript}{manuscript}

\title{Automatic Classification of Defective Photovoltaic Module Cells in
       Electroluminescence Images}
\author[encn,thn,lme]{Sergiu Deitsch\corref{cor1}}
\ead{sergiu.deitsch@fau.de}

\cortext[cor1]{Corresponding author}

\author[lme]{Vincent Christlein}
\author[zae]{Stephan Berger}
\author[zae]{Claudia Buerhop-Lutz}
\author[lme]{Andreas Maier}
\author[encn,thn]{Florian Gallwitz}
\author[lme]{Christian Riess}

\address[encn]{Energie Campus Nuremberg, Fürther~Str.~250, 90429~Nuremberg,
Germany}
\address[thn]{Nuremberg Institute of Technology, Department of Computer Science, Hohfederstr.~40, 90489~Nuremberg, Germany}
\address[zae]{ZAE Bayern, Immerwahrstr.~2, 91058~Erlangen, Germany}
\address[lme]{Pattern Recognition Lab, University of Erlangen-Nuremberg,
  Martensstr.~3, 91058~Erlangen, Germany}

\begin{abstract}
  \Gls{EL} imaging is a useful modality for the inspection of \gls{PV} modules.
  \Gls{EL} images provide high spatial resolution, which makes it possible to
  detect even finest defects on the surface of \gls{PV} modules. However, the
  analysis of \gls{EL} images is typically a manual process that is expensive,
  time-consuming, and requires expert knowledge of many different types of
  defects.

  In this work, we investigate two approaches for automatic detection of such
  defects in a single image of a \gls{PV} cell. The approaches differ in their
  hardware requirements, which are dictated by their respective application
  scenarios. The more hardware-efficient approach is based on hand-crafted
  features that are classified in a \gls{SVM}. To obtain a strong performance,
  we investigate and compare various processing variants. The more
  hardware-demanding approach uses an end-to-end deep \gls{CNN} that runs on a
  \gls{GPU}. Both approaches are trained on \num{1968}~cells extracted from
  high resolution \gls{EL} intensity images of mono- and polycrystalline
  \gls{PV} modules. The \gls{CNN} is more accurate, and reaches an average
  accuracy of~\SI{88.4228187919}{\percent}.  The \gls{SVM} achieves a slightly
  lower average accuracy of~\SI{82.43847875}{\percent}, but can run on arbitrary
  hardware. Both automated approaches make continuous, highly accurate
  monitoring of \gls{PV} cells feasible.
\end{abstract}


\end{frontmatter}

\ifx\arxiv\undefined
\linenumbers
\fi

\section{Introduction}
\label{sec:intro}


\begin{figure*}[tp]
  \setkeys{Gin}{width=(\textwidth-0.75em*4)/5}
  \centering
  \tikzsetnextfilename{solar-cell-defect-examples}%
  \input{figures/defect-examples.tikz}%
  \caption{Various intrinsic and extrinsic defects in monocrystalline
  (\subref{fig:material-defect}--\subref{fig:gridfinger-defect}) and
  polycrystalline (\subref{fig:crack-defect}--\subref{fig:deep-crack-defect})
  solar cells. \subref{fig:material-defect} shows a solar cell with a typical
  material defect. \subref{fig:gridfinger-defect} shows finger interruptions in
  the encircled areas, which do not necessarily reduce the module efficiency.
  The solar cell in~\subref{fig:crack-defect} contains a microcrack that is
  very subtle in its appearance. While microcracks do not divide the cell
  completely, they still must be detected because such cracks may grow over
  time and eventually impair the module efficiency. The spots at the bottom
  of this cell are likely to indicate cell damage as well. However, such spots
  can be oftentimes difficult to distinguish from actual material defects.
  \subref{fig:disconnected-cell-defect} shows a disconnected area due to
  degradation of the cell interconnection. \subref{fig:deep-crack-defect} shows
a cell with electrically separated or degraded parts, which are usually caused
by mechanical damage.}
  \label{fig:pv-module-defects}
\end{figure*}

Solar modules are usually protected by an aluminum frame and glass lamination
from environmental influences such as rain, wind, and snow. However, these
protective measures can not always prevent mechanical damages caused by dropping
the \gls{PV}  module during installation, impact from falling tree branches,
hail, or thermal stress. Also, manufacturing errors such as faulty soldering or
defective wires can also result in damaged \gls{PV} modules. Defects can in turn
decrease the power efficiency of solar modules. Therefore, it is necessary to
monitor the condition of solar modules, and replace or repair defective units in
order to ensure maximum efficiency of solar power plants.

Visual identification of defective units is particularly difficult, even for
trained experts. Aside from obvious cracks in the glass, many defects that
reduce the efficiency of a \gls{PV} module are not visible to the eye.
Conversely, defects that are visible do not necessarily reduce the module
efficiency.

To precisely determine the module efficiency, the electrical output of a module
must be measured directly. However, such measurements require manual
interaction with individual units for diagnosis, and hence they do not scale
well to large solar power plants with thousands of \gls{PV} modules.
Additionally, such measurements only capture one point in time, and as such may
not reveal certain types of small cracks, which will become an issue over
time~\cite{Kajari-Schroder2012}.

\Gls{IR} imaging is a non-destructive, contactless alternative to direct
measurements for assessing the quality of solar modules. Damaged solar modules
can be easily identified by solar cells which are either partially or completely
cut off from the electric circuit. As a result, the solar energy is not
converted into electricity anymore, which heats the solar cells up. The emitted
infrared radiation can then be imaged by an \gls{IR} camera. However, \gls{IR}
cameras are limited by their relatively low resolution, which can prohibit
detection of small defects such as microcracks not yet affecting the
photoelectric conversion efficiency of a solar module.


\Glsfirst{EL} imaging~\cite{Fuyuki2005,Fuyuki2009} is another established
non-destructive technology for failure analysis of \gls{PV} modules with the
ability to image solar modules at a much higher resolution.  In \gls{EL} images,
defective cells appear darker, because disconnected parts do not irradiate. To
obtain an \gls{EL} image, current is applied to a \gls{PV} module, which induces
\gls{EL} emission at a wavelength of \SI{1150}{\nano\meter}. The emission can be
imaged by a silicon \gls{CCD} sensor. The high spatial image resolution enables
the detection of microcracks~\cite{Breitenstein2011}, and \gls{EL} imaging also
does not suffer from blurring due to lateral heat propagation. However, visual
inspection of \gls{EL} images is not only time-consuming and expensive, but also
requires trained specialists. In this work, we remove this constraint by
proposing an automated method for classifying defects in \gls{EL} images.

In general, defects in solar modules can be classified into two
categories~\cite{Fuyuki2009}:
\begin{enumerate*}
  \item intrinsic deficiencies due to material properties such as crystal grain
    boundaries and dislocations, and
  \item process-induced extrinsic defects such as microcracks and breaks, which
    reduce the overall module efficiency over time.
\end{enumerate*}

\Cref{fig:pv-module-defects} shows an example \gls{EL} image with different
types of defects in monocrystalline and polycrystalline solar cells.
\Cref{fig:material-defect} and \cref{fig:gridfinger-defect} show general
material defects from the production process such as finger interruptions which
do not necessarily reduce the lifespan of the affected solar panel unless caused
by high strain at the solder joints~\cite{Koentges2014}. Specifically, the
efficiency degradation induced by finger interruptions is a complex interaction
between their size, position, and the number of
interruptions~\cite{DeRose2012,Koentges2014}.
\crefrange{fig:crack-defect}{fig:deep-crack-defect} show microcracks,
degradation of cell-interconnections, and cells with electrically separated or
degraded parts that are well known to reduce the module efficiency. Particularly
the detection of microcracks requires cameras with high spatial resolution.

\begin{figure*}[tb]
  \centering
  \tikzsetnextfilename{annotated-module-example}%
  \includegraphics[width=\linewidth]{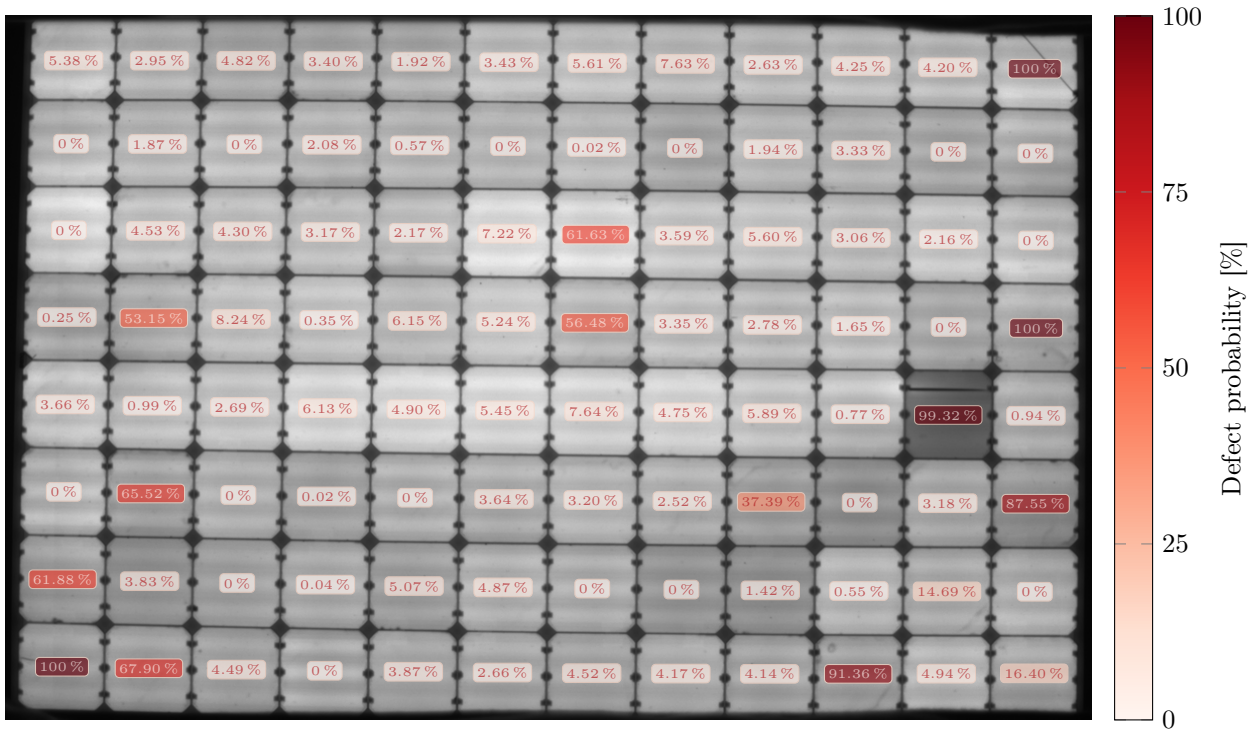}%
  \caption{Defect probabilities inferred for each \gls{PV} module cell by the
  proposed \gls{CNN}. A darker shade of red indicates a higher likelihood of a
  cell defect.}
  \label{fig:el-pv-cell-defect-predictions}
\end{figure*}

For the detection of defects during monitoring one can set different goals.
Highlighting the exact location of defects within a solar module allows to
monitor affected areas with high precision. However, the exact defect location
within the solar cell is less important for the quality assessment of a whole
\gls{PV} module. For this task, the overall likelihood indicating a cell defect
is more important.  \addition[label=a:power-loss,ref=c:power-loss]{This enables
  a quick identification of defective areas and can potentially complement the
prediction of future efficiency loss within a \gls{PV} module.} In this work, we
propose two classification pipelines that automatically solve the second task, i.e.,
to determine a per-cell defect likelihood that may lead to efficiency loss.


The investigated classification approaches in this work are \gls{SVM} and
\gls{CNN} classifiers.
\begin{description}
  \item[Support Vector Machines (SVMs)] are trained on various features
    extracted from \gls{EL} images of solar cells.
  \item[Convolutional Neural Network (CNN)] is directly fed with image pixels of
    solar cells and the corresponding labels.
\end{description}

The \gls{SVM} approach is computationally particularly efficient during training
and inference. This allows to operate the method on a wide range of commodity
hardware, such as tablet computers or drones, whose usage is dictated by the
respective application scenario. Conversely, the prediction accuracy of the
\gls{CNN} is generally higher, while training and inference is much more
time-intensive and commonly requires a \gls{GPU} for an acceptably short
runtime. \addition[label=a:drones,ref=c:drones]{Particularly for aerial imagery,
  however, additional issues may arise and will need to be solved.
\citet{Kang2018} highlight several challenges that need to be addressed before
applying our approach outside of a manufacturing setting}.

\subsection{Contributions}
\label{sec:contribs}

The contribution of this work consists of three parts. First, we present a
resource-efficient framework for supervised classification of defective solar
cells using hand-crafted features and an \Gls{SVM} classifier that can be used
on a wide range of commodity hardware, including tablet computers and drones
equipped with low-power single-board computers. The low computational
requirements make the on-site evaluation of the \gls{EL} imagery possible,
similar to analysis of low resolution \gls{IR} images~\cite{Dotenco2016}.
Second, we present a supervised classification framework using a convolutional
neural network that is slightly more accurate, but requires a \gls{GPU} for
efficient training and classification. In particular, we show how uncertainty
can be incorporated into both frameworks to improve the classification accuracy.
Third, we contribute an annotated dataset consisting of \num{2624}~aligned solar
cells extracted from high resolution \gls{EL} images to the community, and we
use this dataset to perform an extensive evaluation and comparison of the
proposed approaches.

\Cref{fig:el-pv-cell-defect-predictions} exemplarily shows the assessment
results of a solar panel using the proposed convolutional neural network. Each
solar cell in the \gls{EL} image is overlaid by the likelihood of a defect in
the corresponding cell.

\subsection{Outline}

The remainder of this work is organized as follows. Related work is reviewed in
\Cref{sec:pv-cell-related-work}.  \Cref{sec:pv-cell-methodology} introduces
both proposed classification approaches. In \cref{sec:pv-cell-evaluation}, we
evaluate and compare these approaches, and discuss the results.
This work is concluded in \cref{sec:pv-cell-conclusions}.

\section{Related Work}
\label{sec:pv-cell-related-work}

\begin{figure*}[tp]
  \centering
  \tikzsetnextfilename{svm-pipeline}%
  \includegraphics[width=\linewidth]{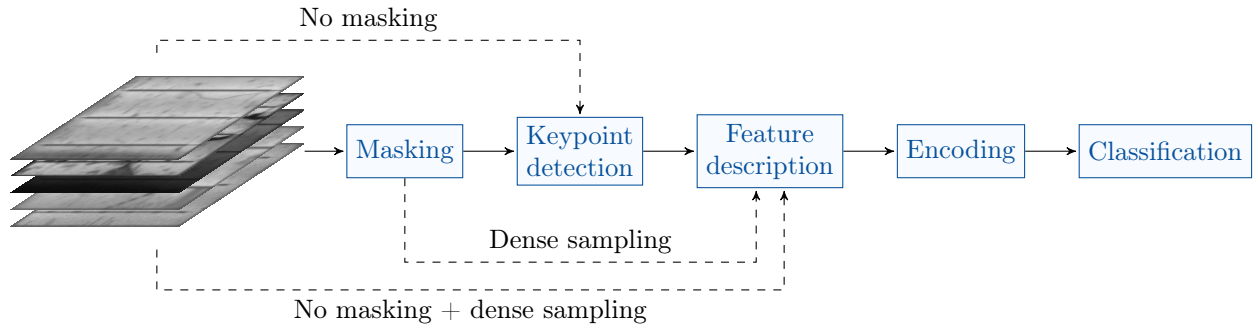}%
  \caption{An overview of the \gls{SVM} classification pipeline with the four
  proposed variations of the preprocessing and feature extraction process.}
  \label{fig:svm-pipeline}
\end{figure*}

Visual inspection of solar modules via \gls{EL} imaging is an active research
topic. Most of the related work, however, focuses on the detection of specific
intrinsic or extrinsic defects, but not on the prediction of defects that
eventually lower the power efficiency of solar modules.
\addition[label=a:health-monitoring,ref=c:health-monitoring]{Detection of
  surface abnormalities in \gls{EL} images of solar cells is related to
  structural health monitoring. However, it is important to note that certain
  defects in solar cells are only specific to \gls{EL} imaging of \gls{PV} modules.
	For instance, fully disconnected solar cells simply appear as dark image
  regions (similar to \cref{fig:disconnected-cell-defect}) and thus have no
  comparable equivalent in terms of structural defects. Additionally, surface
  irregularities in solar wafers (such as finger interruptions) are easily
  confused with cell cracks, even though they do not significantly affect the power
loss}.

In the context of visual inspection of solar modules, \citet{Tsai2012} use
Fourier image reconstruction to detect defective solar cells in \gls{EL} images
of polycrystalline \gls{PV} modules. The targeted extrinsic defects are (small)
cracks, breaks, and finger interruptions. Fourier image reconstruction is
applied to remove possible defects by setting high-frequency coefficients
associated with line- and bar-shaped artifacts to zero. The spectral
representation is then transformed back into the spatial domain. The defects can
then be identified as intensity differences between the original and the
high-pass filtered image. Due to the shape assumption, the method has
difficulties detecting defects with more complex shapes.

\citet{Tsai2013} also introduced a supervised learning method for identification
of defects using \gls{ICA} basis images. Defect-free solar cell subimages are
used to find a set of independent basis images with \gls{ICA}. The method
achieves a high accuracy of~\SI{93.4}{\percent} with a relatively small training
dataset of~\num{300} solar cell subimages. However, material defects such as
finger interruptions are treated equally to cell cracks. This strategy is
therefore only suitable for detection of every abnormality on the surface of
solar cells, but not for the prediction of future energy loss.

\citet{Anwar2014} developed an algorithm for the detection of microcracks in
polycrystalline solar cells. They use anisotropic diffusion filtering followed
by shape analysis to localize the defects in solar cells. While the method
performs well at detecting microcracks, it does not consider other defect types
such as completely disconnected cells, which appear completely dark in \gls{EL}
images.

\citet{Tseng2015} proposed a method for automatic detection of finger
interruptions in monocrystalline solar cells. The method employs binary
clustering of features from candidate regions for the detection of defects.
Finger interruptions, however, do not necessarily provide suitable cues for
prediction of future power loss.


The success of deep learning led to a gradual replacement of traditional pattern
recognition pipelines for optical inspection. However, to our knowledge, no CNN
architecture has been proposed for EL images, but only for other modalities or
applications. Most closely related is the work by \citet{Mehta2017}, who
presented a system for predicting the power loss, localization and type of
soiling from RGB images of solar modules. Their approach does not require manual
localization labels, but instead operates on images with the corresponding power
loss as input. \citet{Masci2012} proposed an end-to-end max-pooling \gls{CNN}
for classifying steel defects. Their network performance is compared against
multiple hand-crafted feature descriptors that are trained using \glspl{SVM}.
Although their dataset consists of only \num{2281} training and \num{646} test
images, the \gls{CNN} architecture classifies steel defects at least twice as
accurately as the \glspl{SVM}. \citet{Zhang2016} proposed a \gls{CNN}
architecture for detection of cracks on roads. To train the \gls{CNN},
approximately \num{45000} hand-labeled image patches were used. They show that
\glspl{CNN} greatly outperform hand-crafted features classified by a combination
of an \gls{SVM} and boosting.
\addition[label=a:more-related-work,ref=c:health-monitoring]{\citet{Cha2017} use
  a very similar approach to detect concrete cracks in a broad range of images
taken under various environmental and illumination conditions.}
\addition[label=a:drones-ref,ref=c:drones]{\citet{Kang2018} employ deep learning for
structural health monitoring on aerial imagery.}
\addition[label=a:flexible-bb,ref=c:bb-size]{\citet{Cha2018} additionally investigated
defect localization using the modern learning-based segmentation approaches for
region proposals based on the Faster R-CNN framework which can perform in
real-time.} \citet{Lee2019} also use semantic segmentation to detect cracks in
concrete.

In medical context, \citet{Esteva2017} employ deep neural networks to classify
different types of skin cancer. They trained the \gls{CNN} end-to-end on a large
dataset consisting of \num{129450} clinical images and \num{2032} different
diseases making it possible to achieve a high degree of accuracy.

\section{Methodology}
\label{sec:pv-cell-methodology}

We subdivide each module into its solar cells, and analyze each cell
individually to eventually infer the defect likelihood. This breaks down the analysis to the smallest meaningful unit, in
the sense that the mechanical design of \gls{PV} modules interconnects units of
cells in series. Also, the breakdown considerably increases the number of
available data samples for training. For the segmentation of solar cells, we use
a recently developed method~\cite{Deitsch2018}, which brings every cell into a
normal form free of perspective and lens distortions.

\addition[label=a:window-size,ref=c:bb-size]{Unless otherwise stated, the
  proposed methods operate on size-normalized \gls{EL} images of solar cells
  with a resolution of~\(300\times300\) pixels. This image resolution was
  derived from the median dimensions of image regions corresponding to
  individual solar cells in the original \gls{EL} images of \gls{PV} modules.
  The solar cell images are used directly as pipeline input.
  The image resolution of solar cells in the wild will generally deviate from
  the required resolution and therefore must be adjusted accordingly.
  The \gls{CNN} architecture sets a minimum image resolution, which typically
  equals the \gls{CNN}'s receptive field (\eg, the original \vggnet\
  architecture uses \(224\times224\)).  If the resolution is lower than this
  minimum resolution, then the image must be upscaled.  For higher resolutions,
  the network can be applied in a strided window manner and afterwards the
  outputs are pooled together (typically using average or maximum
  pooling). We followed an alternative approach in which the \gls{CNN}
  architecture encodes this process inherently. In case of the \gls{SVM}
  pipeline, the resolution requirement is less stringent. Given local features
  that are scale-invariant, the image resolution of the classified solar cells
  does not need to be adjusted and may vary from image to image.}

\subsection{Classification Using Support Vector Machines}

The general approach for classification using \glspl{SVM}~\cite{Cortes1995} is
as follows. First, local descriptors are extracted from images of segmented
\gls{PV} cells. The features are typically extracted at salient points, also
known as \emph{keypoints}, or from a dense pixel grid. For training
the classifier and subsequent predictions, a global representation needs to be
computed from the set of local descriptors, oftentimes referred to as
\emph{encoding}. Finally, this global descriptor for a solar cell is classified
into  defective or functional. \Cref{fig:svm-pipeline} visualizes the
classification pipeline, consisting of masking, keypoint detection, feature
description, encoding, and classification. We describe these steps in the
following subsections.

\subsubsection{Masking}\label{sec:masking}

We assume that the solar cells were segmented from a \gls{PV} module, \eg,
using the automated algorithm we proposed in earlier work~\cite{Deitsch2018}. A binary mask
allows then to separate the foreground of every cell from the background. The
cell background includes image regions that generally do not belong to the
silicon wafer, such as the busbars and the inter-cell borders. This mask can be
used to strictly limit feature extraction to the cell interior. In the
evaluation, we investigate the usefulness of masking, and find that its effect
is minor, \ie, it only slightly improves the performance in a few
feature/classifier combinations.

\subsubsection{Feature Extraction}
\label{sec:feature-extraction}

\begin{figure*}[tb]
  \centering
  \tikzsetnextfilename{keypoint-detection-strategies}%
  \input{figures/keypoint-detection-strategies}%
  \caption{Two different feature extraction strategies applied to the same
    \gls{PV} cell with and without masking.
    In~\subref{fig:pv-cell-dense-sampling}, keypoints are sampled at fixed
    positions specified by the center of a cell in the overlaid grid.
    \subref{fig:pv-cell-masked-dense-points} uses equally sized and oriented
    keypoints laid out on a dense grid similar to
    \subref{fig:pv-cell-dense-sampling}. \subref{fig:pv-cell-keypoint-detection}
    shows an example for AGAST keypoints (detection threshold slightly increased
    for visualization). \subref{fig:pv-cell-masked-keypoint-detection} shows
  KAZE keypoints of various sizes and orientations after masking out the
background area.}
  \label{fig:pv-cell-feature-detection}
\end{figure*}

In order to train the \glspl{SVM}, feature descriptors are extracted first. The
locations of these local features are determined using two main sampling
strategies:
\begin{enumerate*}[label=(\arabic*)]
  \item keypoint detection, and
  \item dense sampling.
\end{enumerate*}
These strategies are exemplarily illustrated in
\cref{fig:pv-cell-feature-detection}. \addition[label=a:kps1,ref=c:kps]{Both
  strategies produce different sets of features that can be better suitable for
  specific types of solar wafers than others. Dense sampling disregards the
  image content and instead uses a fixed configuration of feature points.
  Keypoint detectors, on the other hand, rely on the textureness in the image
  and therefore the number of keypoints is proportional to the amount of
high-frequency elements, such as edges and corners (as can be seen in
\cref{fig:pv-cell-keypoint-detection,fig:pv-cell-masked-keypoint-detection}).
} Keypoint detectors typically operate in scale space, allowing feature
detection at different scale levels and also at different orientations.
\addition[label=a:kps2,ref=c:kps]{\Cref{fig:pv-cell-masked-keypoint-detection}
shows keypoints detected by KAZE. Here, each keypoint has a different scale
(visualized by the radius of corresponding circles) and also a specific
orientation exemplified by the line drawn from the center to the circle border.
Keypoints that capture both the scale and the rotation are invariant to changes
in image resolution and to in-plane rotations, which makes them very robust.}

Dense sampling subdivides the \(300\times300\)~pixels \gls{PV} cell by
overlaying it with a grid consisting of \(n\times n\)~cells. The center of each
grid cell specifies the position at which a feature descriptor will be
subsequently extracted.  The number of feature locations only depends on the
grid size. \addition[label=a:kps3,ref=c:kps]{Dense sampling can be useful if
computational resources are very limited, or if the purpose is to identify
defects only in monocrystalline \gls{PV} modules.}

We employ different popular combinations of keypoint detectors and feature
extractors from the literature, as listed in
\cref{tab:investigated_detectors_and_descriptors} and outlined below.

\begin{table}[tb]
\caption{Investigated keypoint detectors and feature descriptors. SIFT, SURF,
and KAZE (in bold) contain both a detector and a descriptor. We explored
also combinations of the keypoint detectors of AGAST and KAZE with other
feature descriptors. Note, the keypoints provided by SIFT and SURF were not reliable
enough and thus not further evaluated.}
\label{tab:investigated_detectors_and_descriptors}
\centering
\begin{tabularx}{\columnwidth}{r|YY}
  \toprule
  \multicolumn{1}{c|}{Method}
  & Keypoint detector & Feature descriptor
  \\
  \midrule
  AGAST~\cite{Mair2010} & \cmark & \xmark
  \\
  \textbf{KAZE}~\cite{Alcantarilla2012} & \cmark & \cmark
  \\
  HOG~\cite{Dalal2005} & \xmark & \cmark
  \\
  PHOW~\cite{Bosch2007} & \xmark & \cmark
  \\
  \textbf{SIFT}~\cite{Lowe1999} & (\cmark) & \cmark
  \\
  \textbf{SURF}~\cite{Bay2008} & (\cmark) & \cmark
  \\
  VGG~\cite{Simonyan2014a} & \xmark & \cmark
  \\
  \bottomrule
\end{tabularx}
\end{table}

Several algorithms combine keypoint detection and feature description. Probably
the most popular of these methods is \gls{SIFT}~\cite{Lowe1999}, which detects
and describes features at multiple scales. \gls{SIFT} is invariant to rotation,
translation, and scaling, and partially resilient to varying illumination
conditions. \gls{SURF}~\cite{Bay2008} is a faster variant of \gls{SIFT}, and
also consists of a keypoint detector and a local feature descriptor.  However,
the detector part of \Gls{SURF} is not invariant to affine transformations. In
initial experiments, we were not able to successfully use the keypoint
detectors of \gls{SIFT} and \gls{SURF}, because the keypoint detector at times
failed to detect features in relatively homogeneous monocrystalline cell
images, and hence we used only the descriptor parts.

KAZE~\cite{Alcantarilla2012} is a multiscale feature detector and descriptor.
The keypoint detection algorithm is very similar to \gls{SIFT}, except that the
linear Gaussian scale space used by \gls{SIFT} is replaced by nonlinear
diffusion filtering. For feature description, however, KAZE uses the \gls{SURF}
descriptor.

We also investigated \Glsfirst{AGAST}~\cite{Mair2010} as a dedicated keypoint
detector without descriptor. It is based on a random forest classifier trained
on a set of corner features that is known as
\gls{FAST}~\cite{Rosten2005,Rosten2006}.

Among the dedicated descriptors, \Gls{PHOW}~\cite{Bosch2007} is an extension of
\gls{SIFT} that computes \gls{SIFT} descriptors densely over a uniformly spaced
grid. We use the implementation variant
from~\textsc{Vlfeat}~\cite{Vedaldi2008}.  Similarly, \Gls{HOG}~\cite{Dalal2005}
is a gradient-based feature descriptor computed densely over a uniform set of
image blocks.  Finally, we also used the \gls{VGG} descriptor trained
end-to-end using an efficient optimization method~\cite{Simonyan2014a}. In our
implementation, we employ the 120-dimensional real-valued descriptor variant.

We omitted binary descriptors from this selection. Even though binary feature
descriptors are typically very fast to compute, they generally do not perform
better than real-valued descriptors~\cite{Heinly2012}.

\subsubsection{Combinations of Detectors and Extractors}

For the purpose of determining the most powerful feature detector/extractor
combination, we evaluated all  feature detector and feature extractor
combinations with few exceptions.

In most cases, we neither tuned the parameters of keypoint detectors  nor those
of feature extractors but rather used the defaults by \textsc{Opencv}~\cite{Itseez2017}
as of version~3.3.1. One notable exception is \gls{AGAST}, where we lowered the
detection threshold to~5 to be able to detect keypoints in
monocrystalline \gls{PV} modules. For \gls{SIFT} and \gls{SURF}, similar
adjustments were not successful, which is why we only used their descriptors.
\gls{HOG} requires a grid of overlapping image regions, which is incompatible
with the keypoint detectors. Instead, we downsampled the \(300\times300\)
pixels cell images to \( 256 \times 256 \)~pixels (the closest power of~2) for
feature extraction. Masking was omitted for \gls{HOG} due to
implementation-specific limitations.
Given these exceptions, we overall evaluate twelve feature
combinations.

\subsubsection{Encoding}

The computed features are encoded into a global feature descriptor.
The purpose of encoding is the formation of a single, fixed-length global
descriptor from multiple local descriptors. Encoding is commonly represented as
a histogram that draws its statistics from a background model.
To this end, we employ \gls{VLAD}~\cite{Jegou12ALI}, which offers a compact
state-of-the-art representation~\cite{Peng15}. VLAD encoding is sometimes also
used for deep learning based features in classification, identification and
retrieval tasks~\cite{Gong14MSO,Ng15,Paulin16,Christlein2017}.

The \gls{VLAD} dictionary is created by \(k\)-means clustering of a random
subset of feature descriptors from the training set. For performance reasons,
we use the fast mini-batch variant~\cite{Sculley10} of \(k\)-means.
The cluster centroids~\(\vec\mu_k\) correspond to
anchor points of the dictionary. Afterwards,
first order statistics are aggregated as a sum of
residuals of all descriptors~\(\mathcal{X} \coloneqq\{\bx_t \in \mathbb{R}^d
\mid t=1,\dotsc, T\}\) extracted from a solar cell image. The residuals are
computed with respect to their nearest anchor point~\(\bmu_k\) in the
dictionary~\(D\coloneqq\{\bmu_k \in \mathbb{R}^d \mid k=1,\dotsc,K\}\) as
\begin{gather}
\label{eq:vlad-residual}
  \vec{\nu}_{k}
  \coloneqq \sum_{t=1}^T \eta_k(\bx_t) (\bx_t - \bmu_k)
  \\
  \intertext{where \( \eta_k \colon \mathbb{R}^d \to \{0,1\} \) is an indicator
function to cluster membership, \ie,}
 \eta_{k}(\bx) \coloneqq
  \begin{dcases}
    1 &\text{if}~k=\argmin_{j=1,\dotsc,K} \lVert \bx - \bmu_j \rVert_2\\
    0 &\text{otherwise}
  \end{dcases}
  \enspace,
\end{gather}
which indicates whether~\( \vec x \) is the nearest neighbor of \(\vec\mu_k \).
The final \gls{VLAD} representation~\(\vec{\nu} \in \mathbb{R}^{Kd}\) corresponds
to the concatenation of all residual terms~\eqref{eq:vlad-residual} into
a~\(Kd\)-dimensional vector:
\begin{equation}
  \vec{\nu}\coloneqq(\vec{\nu}^{\top}_1,\ldots,\vec{\nu}^{\top}_K)^{\top}
  \enspace.
\end{equation}

Several normalization steps are required to make the \gls{VLAD} descriptor robust.
Power normalization addresses issues when some local descriptors occur more
frequently than others. Here, each element of the global descriptor~\(v_i \in
\vec{\nu}\) is normalized as
\begin{equation}
  \hat{v}_i \coloneqq \sign(v_i) \abs{v_i}^{\rho},
  \quad
  i=1,\dotsc,Kd
\end{equation}
where we chose~\(\rho=0.5\) as a typical value from the literature.
After power normalization, the vector is normalized such that its
\(\ell^2\)-norm equals one.

\begin{figure*}[!tbp]
  \centering
  \tikzsetnextfilename{vgg19-architecture}%
  \tikzpicturedependsonfile{vgg19styles.tex}%
  \includegraphics[width=\linewidth]{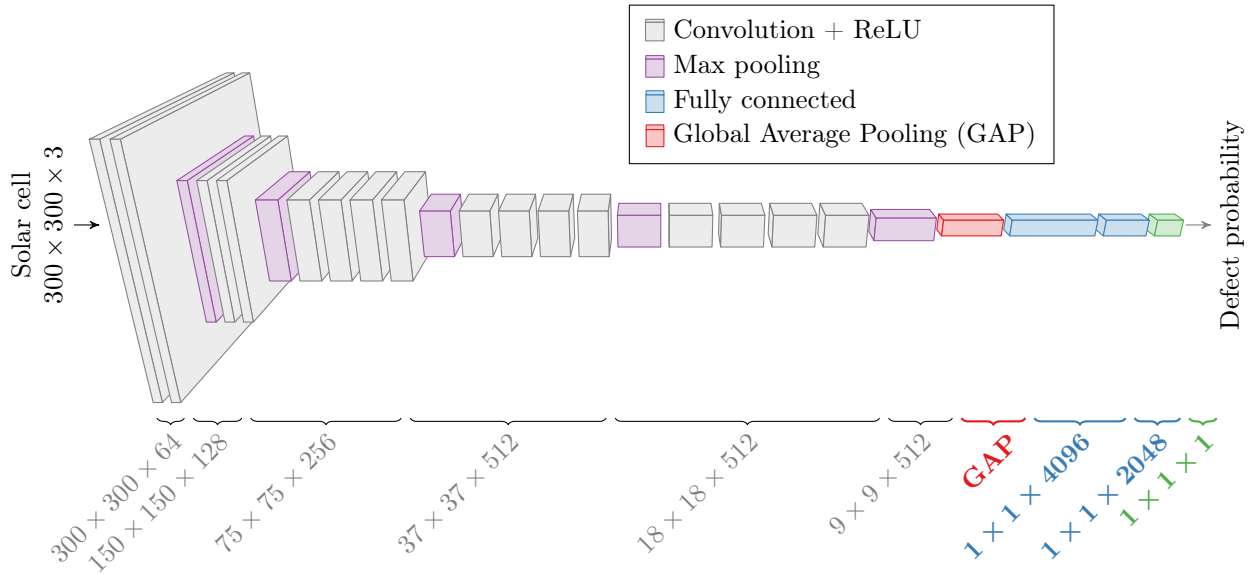}%
  \caption{Architecture of the modified \vggnet\ network used for prediction of
  defect probability in \(300\times300\)~pixels \gls{EL} images of solar cells.
  Boldface denotes layers that deviate from \vggnet.}
  \label{fig:vgg19}
\end{figure*}

Similarly, an over-counting of \emph{co-occurrences} can occur if at least two
descriptors appear together frequently. \citet{Jegou12NEA} showed that \gls{PCA}
whitening effectively eliminates such co-occurrences and additionally
decorrelates the data.

To enhance the robustness of the codebook~\(D\) against potentially suboptimal
solutions from the probabilistic \(k\)-means clustering, we compute five
\gls{VLAD} representations from different training subsets using different
random seeds. Afterwards, the concatenation of the \gls{VLAD}
encodings~\(\tilde{\vec{\nu}}\coloneqq
(\hat{\vec{\nu}}^\top_1,\ldots,\hat{\vec{\nu}}^\top_m)^{\top} \in
\mathbb{R}^{mKd} \) is jointly decorrelated and whitened by means of
\gls{PCA}~\cite{Kessy2018}. The transformed representation is again normalized
such that its \(\ell^2\)-norm equals one and the result is eventually passed
to the \gls{SVM} classifier.

\subsubsection{Support Vector Machine Training}

We trained \glspl{SVM} both with a linear and a \gls{RBF} kernel. For the
linear kernel, we use \textsc{Liblinear}~\cite{Fan2008},
which is optimized for linear classification tasks and large datasets.
For the non-linear \gls{RBF} kernel, we use \textsc{Libsvm}~\cite{Chang2011}.

The \gls{SVM} hyperparameters are determined by evaluating the average \( F_1
\)~score~\cite{Rijsbergen1979} in an inner five-fold cross-validation on the
training set using a grid search. For the linear \gls{SVM}, we employ the
\(\ell^2\) penalty on a squared hinge loss. The penalty parameter~\(C\) is
selected from a set of powers of ten, \ie, \( C_{\text{linear}}\in \{ 10^k \mid
k=-2,\dotsc,6 \}\subset \mathbb{R}_{>0}\). For \gls{RBF} \glspl{SVM}, the
penalty parameter~\(C\) is determined from a slightly smaller set
\(C_{\text{RBF}}\in \{ 10^k \mid k=2,\dotsc,6\}\). The search space of the
kernel coefficient~\(\gamma\) is constrained to \(\gamma\in\{10^{-7},10^{-6},
S^{-1} \} \subset [0, 1]\), where \(S\)~denotes the number of training samples.

\subsection{Regression Using a Deep Convolutional Neural Network}

We considered several strategies to train the \gls{CNN}. Given the limited
amount of data we had at our disposal, best results were achieved by means of
transfer learning. We utilized the \vggnet\ network
architecture~\cite{Simonyan2015} originally trained on the \textsc{ImageNet}
dataset~\cite{Deng2009} using \num{1.28} million images and \num{1000}~classes.
We then refined the network using our dataset.

We replaced the two fully connected layers of~\vggnet\ by a
\gls{GAP}~\cite{Lin2013} and two fully connected layers with \num{4096} and
\num{2048}~neurons, respectively (\cf, \cref{fig:vgg19}). The \gls{GAP} layer is
used to make the \vggnet\ network input tensor (\(224\times224\times3\))
compatible to the resolution of our solar cell image samples
(\(300\times300\times3\)), in order to avoid additional downsampling of the
samples.
The output layer consists of a single neuron that outputs the defect probability
of a cell. The \gls{CNN} is refined by minimizing the \gls{MSE} loss function.
Hereby, we essentially train a deep regression network, which allows us to
predict (continuous) defect probabilities trained using only two defect
likelihood categories (functional and defective). By rounding the predicted
continuous probability to the nearest neighbor of the four original classes, we
can directly compare \gls{CNN} decisions against the original ground truth
labels without binarizing them.

Data augmentation is used to generate additional, slightly perturbed training
samples. The augmentation variability, however, is kept moderate, since the
segmented cells vary only by few pixels along the translational axes, and few
degrees along the axis of rotation. The training samples are scaled by at
most~\SI{2}{\percent} of the original resolution. The rotation range is capped
to~\(\pm\SI{3}{\degree}\). The translation is limited
to~\(\pm\SI{2}{\percent}\) of the cell dimensions. We also use random flips
along the vertical and horizontal axes. Since the busbars can be laid out both
vertically and horizontally, we additionally include training samples rotated
by exactly~\SI{90}{\degree}. The rotated samples are augmented the same way as
described above.

We fine-tune the pretrained \textsc{ImageNet} model on our data to adapt the
\gls{CNN} to the new task, similar to \citet{Girshick2014}. We, however, do this
in two stages. First, we train only the fully connected layers with randomly
initialized weights while keeping the weights of the convolutional blocks fixed.
Here, we employ the \textsc{Adam} optimizer~\cite{Kingma2014} with a learning
rate of~\(10^{-3}\), exponential decay rates \(\beta_1=0.9\) and
\(\beta_2=0.999\), and the regularization value~\(\hat{\epsilon}=10^{-8}\). In
the second step, we refine the weights of all layers. At this stage, we use the
\gls{SGD} optimizer with a learning rate of~\(5\cdot 10^{-4}\) and a momentum
of~0.9. \addition[label=a:fine-tune,ref=c:two-stage]{We observed that
fine-tuning the \gls{CNN} in several stages by subsequently increasing the
number of hyperparameters slightly improves the generalization ability of the
resulting model compared to a single refinement step.}

In both stages, we process the augmented versions of the \num{1968}~training
samples in mini-batches of 16~samples on two NVIDIA GeForce GTX~1080, and run
the training procedure for a maximum of \num{100}~epochs. This totals to
\num{196800}~augmented variations of the original \num{1968}~training samples
that are used to refine the network. For the implementation of the deep
regression network, we use \textsc{Keras} version
2.0~\cite{Chollet2015} with \textsc{TensorFlow}
version 1.4~\cite{Abadi2015} in the backend.

\section{Evaluation}
\label{sec:pv-cell-evaluation}

For the quantitative evaluation, we first evaluate different feature descriptors
extracted densely over a grid. Then, we compare the best configurations against
feature descriptors extracted at automatically detected keypoints to determine
the best performing variation of the \gls{SVM} classification pipeline. Finally,
we compare the latter against the proposed deep \gls{CNN}, and visualize the
internal feature mapping of the \gls{CNN}.

\subsection{Dataset}
\label{sec:dataset}

We propose a public dataset\footnote{The solar cell dataset is available
at~\url{https://github.com/zae-bayern/elpv-dataset}} of solar cells extracted
from high resolution \gls{EL} images of monocrystalline and polycrystalline
\gls{PV} modules~\cite{Buerhop2018}. The dataset consists of \num{2624}~solar
cell images at a resolution of \(300\times300\)~pixels originally extracted from
\num{44}~different \gls{PV} modules, where \num{18}~modules are of
monocrystalline type, and \num{26}~are of polycrystalline type.

\addition[label=a:conditions,ref=c:conditions]{The images of \gls{PV} modules
used to extract the individual solar cell samples were taken in a manufacturing
setting. Such controlled conditions enable a certain degree of control on
quality of imaged panels and allow to minimize negative effects on image
quality, such as overexposure. Controlled conditions are also required
particularly because background irradiation can predominate \gls{EL}
irradiation. Given \gls{PV} modules emit the only light during acquisition
performed in a dark room, it can be ensured the images are uniformly
illuminated. This is opposed to image acquisition in general structural health
monitoring, which introduces additional degrees of freedom where images can
suffer from shadows or spot lighting~\cite{Cha2017}. An important issue in
\gls{EL} imaging, however, can be considered blurry (\ie, out-of-focus) \gls{EL}
images due to incorrectly focused lens which can be at times challenging to
attain. Therefore, we ensured to include such images in the proposed dataset
(\cf, \cref{fig:pv-module-defects} for an example).}

The solar cells
exhibit intrinsic and extrinsic defects commonly occurring in mono- and
polycrystalline solar modules. In particular, the dataset includes microcracks
and cells with electrically separated and degraded parts, short-circuited cells,
open inter-connects, and soldering failures. These cell defects are widely known
to negatively influence efficiency, reliability, and durability of solar
modules. Finger interruptions are excluded since the power loss caused by such
defects is typically negligible.

\begin{table}[tb]
\caption{Partitioning of the solar cells into functional and defective, with an
additional self-assessment on the rater's confidence after visual inspection.
Non-confident decisions obtain a weight
lower than~\SI{100}{\percent} for the evaluation of the classifier performance.}
\label{fig:pv-cell-defect-prob-tree}
\begin{tabular}{rc|cr}
\toprule
Condition & Confident? & Label \(p\) & Weight \(w\) \\
\midrule
\multirow{2}{*}{functional} & \cmark & functional & \SI{100}{\percent}
\\
& \xmark & defective  & \SI{33}{\percent}
\\
\midrule
\multirow{2}{*}{defective}  & \cmark & defective  & \SI{100}{\percent}
\\
& \xmark & defective  & \SI{67}{\percent}
\\
\bottomrule
\end{tabular}
\end{table}

Measurements of power degradation were not available to provide the ground
truth. Instead, the extracted cells were presented in random order to a
recognized expert, who is familiar with intricate details of different defects
in \gls{EL} images. The criteria for such failures are summarized by
\citet{Koentges2014}. In their failure categorization, the expert focused
specifically on defects with known power loss above~\SI{3}{\percent} from the
initial power output. The expert answered the questions
\begin{enumerate*}[label=(\arabic*)]
  \item is the cell \emph{functional} or \emph{defective}?
  \item are you \emph{confident} in your assessment?
\end{enumerate*}
The assessments into functional and defective cells by a confident rater were
directly used as labels. Non-confident assessments of functional and defective
cells were all labeled as defective. To reflect the rater's uncertainty, lower
weights are assigned to these assessments, namely a weight of~\SI{33}{\percent}
to a non-confident assessment of functional cell, and a weight
of~\SI{67}{\percent} to a non-confident assessment of defective cell.
\Cref{fig:pv-cell-defect-prob-tree} shows this in summary, with the rater
assessment on the left, and the associated classification labels and weights on
the right.  \Cref{tab:pv-cell-labels-distribution} shows the distribution of
ground truth solar cell labels, separated by the type of the source \gls{PV}
module.

\begin{table*}[!bp]
\centering
\caption{The distribution of the total number of solar cell images in the
  dataset depending on sample label~\(p\) and the \gls{PV} module type from which
  the solar cells were originally extracted. The numbers of solar cell images
  are given for the \SI{75}{\percent}/\SI{25}{\percent} training/test split.}
\label{tab:pv-cell-labels-distribution}
\pgfplotstableset{search path={data}, col sep=comma}%
\pgfplotstableread{dataset_stats.csv}\stats%
\pgfplotstabletypeset
[
  column type=r,
  percent/.style={
    column name=\SI{#1}{\percent},
  },
  every head row/.style={
    before row={
      \toprule
        \multirow{2}{*}{Solar wafer}%
      & \multicolumn{4}{c|}{\textbf{Train}}
      & \multicolumn{4}{c|}{\textbf{Test}}
      & \multirow{2}{*}{\(\Sigma\)}%
      \\
    },
    after row=\midrule,
  },
  every last row/.style={
    before row=\midrule,
    after row=\bottomrule,
  },
  multirow header/.style={
    assign cell content/.code={%
      \ifnum\pgfplotstablerow=0
        \pgfkeyssetvalue{/pgfplots/table/@cell content}%
          {\multirow{#1}{*}{A##1}}%
      \else
        \pgfkeyssetvalue{/pgfplots/table/@cell content}{}%
      \fi
    },
  },
  columns/type/.style={
    string replace={mono}{Monocrystalline},
    string replace={poly}{Polycrystalline},
    string replace={overall}{\(\Sigma\)},
    column type=r,
    column name=,
    string type,
  },
  columns/0 train/.style={
    percent=0,
    column type/.add={|}{},
  },
  columns/33 train/.style={
    percent=33,
  },
  columns/67 train/.style={
    percent=67,
  },
  columns/100 train/.style={
    percent=100,
    column type/.add={}{|},
  },
  columns/0 test/.style={
    percent=0,
  },
  columns/33 test/.style={
    percent=33,
  },
  columns/67 test/.style={
    percent=67,
  },
  columns/100 test/.style={
    percent=100,
    column type/.add={}{|},
  },
  columns/overall/.style={
    column type=c,
    column name=,
  }
]\stats
\end{table*}

We used \SI{25}{\percent} of the labeled cells (\num{656}~cells)
for testing, and the remaining \SI{75}{\percent} (\num{1968}~cells) for
training. Stratified sampling was used to randomly split the samples while
retaining the distribution of samples within different classes in the
training and the test sets.
To further balance the training set, we weight the classes using the inverse
proportion heuristic derived from \citet{King2001}
\begin{equation} c_j
\coloneqq
  \frac{S}{2 n_j}
  \enspace ,
\end{equation}
where \(S\)~is the total number of training samples, and \( n_j\)~is the number
of functional (\(j=0\)) or defective (\(j=1\)) samples.

\subsection{Dense Sampling}

\begin{figure*}[!tb]
  \centering
  \tikzsetnextfilename{dense-conf-per-feature}%
  \includegraphics[width=\linewidth]{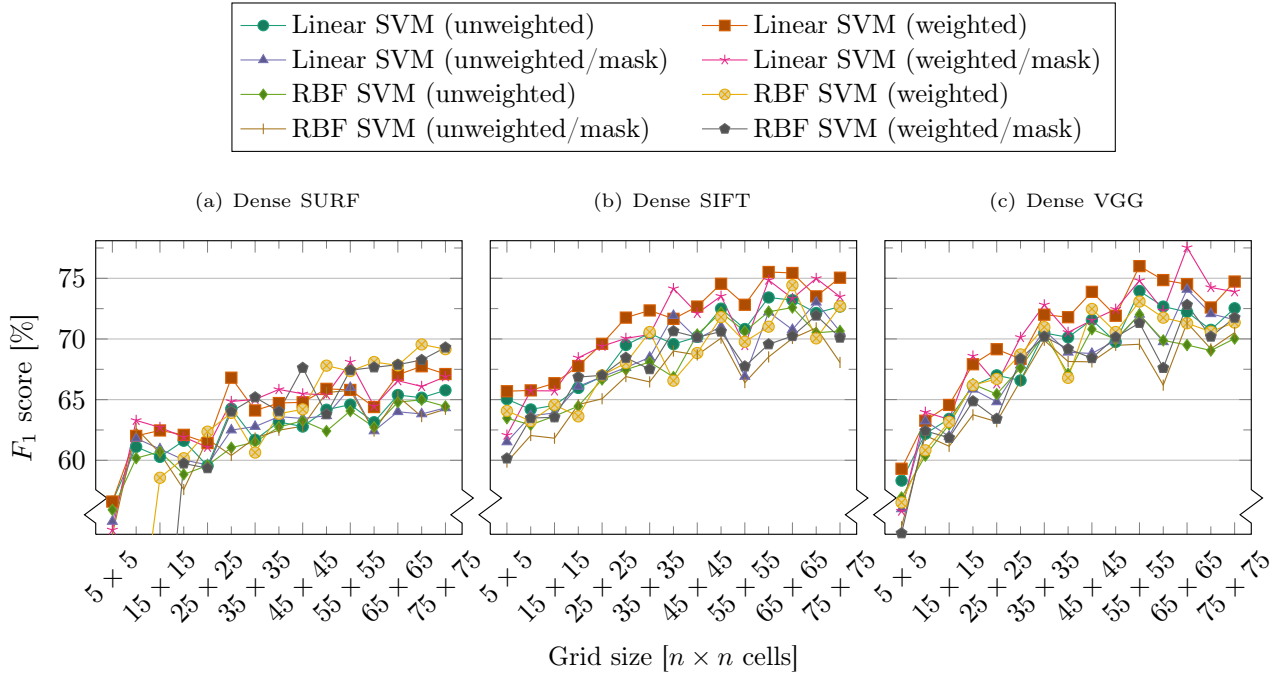}%
  \caption{Classification performance for different dense sampling
  configurations in terms of \(F_1\)~score grouped by the feature descriptor,
  classifier, weighting strategy, and the use of masking. The highest
  \(F_1\)~score is achieved using a linear \gls{SVM} and the \gls{VGG} feature
  descriptor at a grid resolution of \(65\times65\)~cells with sample weighting
  and
  masking~(\ref*{pgfplots:dense-vgg-weighted-masked})~\subref{fig:dense-conf-vgg}.}
  \label{fig:el-pv-cell-dense-conf}
\end{figure*}

In this experiment, we evaluate different grid sizes for subdividing a single
\(300\times 300\)~pixels cell image. The number of grid points per cell is
varied between \(5\times5\) to \(75\times75\) points. At each grid point,
\gls{SIFT}, \gls{SURF}, and \gls{VGG} descriptors are computed. The remaining
two descriptors, \gls{PHOW} and \gls{HOG}, are omitted in this experiment,
because they do not allow to arbitrarily specify the position for descriptor
computation.  Note that at a~\(75\times 75\) point grid, the distance between
two grid points is only 4~pixels, which leads to a significant overlap between
neighbored descriptors.  Therefore, further increase of the grid resolution
cannot be expected to considerably improve the classification results.

The goal of this experiment is to find the best performing combination of grid
size and classifier.  We trained both linear \glspl{SVM} and \glspl{SVM} with
the \gls{RBF} kernel.  For each classifier, we also examine two additional
options, namely whether the addition of the sample weights~\(w\) (\cf\
\cref{fig:pv-cell-defect-prob-tree}) or masking out the background region
(\cf\ \cref{sec:masking}) improves the classifiers.

\begin{figure*}[tb]
  \centering
  \tikzsetnextfilename{roc-overview}
  \includegraphics[width=\linewidth]{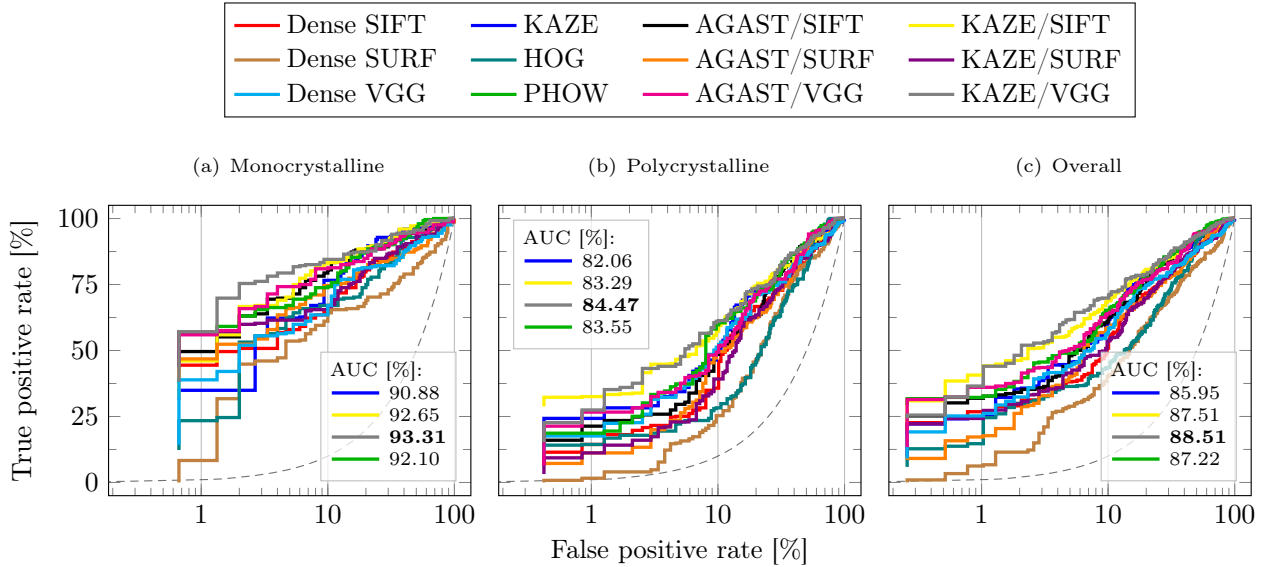}%
  \caption{\Glsfirst{ROC} for top performing feature detector/extractor
    combinations grouped by mono-, polycrystalline, and both solar module types
    combined. The dashed curve~(\rocbaselinemarker) represents the baseline in
    terms of a random classifier. Note the logarithmic scale of the false
  positive rate axis. Refer to the text for details.}
  \label{fig:roc-baseline}
\end{figure*}

Performance is measured using the \(F_1\)~score, which is the harmonic mean of
precision and recall. \Cref{fig:el-pv-cell-dense-conf} shows the
\(F_1\)~scores that are averaged over the individual per-class \(F_1\)~scores.
From left to right, these scores are shown for the \gls{SURF} descriptor
(\cref{fig:dense-conf-surf}), \gls{SIFT} descriptor
(\cref{fig:dense-conf-sift}) and VGG descriptor (\cref{fig:dense-conf-vgg}).
Here, the \gls{VGG} descriptor achieves the highest score on a grid of
size~\(65\times65\) using a linear \gls{SVM} with weighting and masking.
\gls{SIFT} is the second best performing descriptor with best performance on a
\(60\times60\) grid using linear \gls{SVM} with weighting, but without masking.
\gls{SURF} achieved the lowest scores, with a peak at a~\(70\times70\) grid
using an \gls{RBF} \gls{SVM} with weighting, but without masking. The results
show the trend that more grid points lead to better results.  The
classification accuracy of \gls{SURF} increases only slowly and saturates at
about~\SI{70}{\percent}. \gls{SIFT} and \gls{VGG} benefit more from denser
grids. The use of the weights~\(w\) leads in most cases to a higher score,
because the classifier can stronger rely on samples for which the expert
labeler was more confident. Masking also improves the \( F_1 \)~score for VGG
features. However, the improvement by almost two percent is small compared to
the overall performance variation over the configurations. One can argue that
the cell structure is not substantial for distinguishing different kinds of
cell defects given the high density of the feature points and the degree of
overlap between image regions evaluated by feature extractors.

\subsection{Dense Sampling vs.\ Keypoint Detection}

This experiment aims at comparing the classification performance of dense
grid-based features versus keypoint-based features.  To this end, the best
performing grid-based classifier per descriptor from the previous experiment
are compared to combinations of keypoint detectors and feature descriptors.

\begin{figure*}[tbp]
  \centering
  \tikzsetnextfilename{roc-cnn-vs-baseline-comparison}%
  \includegraphics[width=\linewidth]{figures/roc-cnn-vs-baseline-comparison}
  \caption{\Gls{ROC} curves of the best performing KAZE/\gls{VGG} feature
  detector/descriptor combination (\ref*{pgfplots:roc-baseline-mono}) compared
  to the \gls{ROC} of the deep regression network
  (\ref*{pgfplots:roc-cnn-mono}). While in the monocrystalline
  case~\subref{fig:roc-final-mono} the classification performance of the
  \gls{CNN} is almost on par with the linear \gls{SVM}. For polycrystalline
  \gls{PV} modules~\subref{fig:roc-final-poly} the \gls{CNN} considerably
  outperforms \gls{SVM} with the linear kernel trained on KAZE/\gls{VGG}
  features. The latter outcome leads to a higher \gls{CNN} \gls{ROC} \gls{AUC}
  for both \gls{PV} modules types combined~\subref{fig:roc-final-overall}. The
  dashed curve~(\rocbaselinemarker) represents the baseline in terms of a
  random classifier.}
  \label{fig:comparison_svm_cnn}
\end{figure*}

\Cref{fig:roc-baseline} shows the evaluated detector and extractor combinations
for monocrystalline cells, polycrystalline cells, and both together. Most
detector/extractor combinations are specified by a forward
slash~(\textsl{Detector}/\textsl{Descriptor}). Entries
without a forward slash, namely KAZE, \gls{HOG}, and \gls{PHOW}, denote
features which already include both a detector and a descriptor. The three best
performing methods on a dense grid are denoted as Dense SIFT \(60\times 60\),
Dense SURF \(70\times70\), and Dense VGG \(65\times 65\), respectively.  Unless
otherwise specified, the features were trained with sample weighting, without
masking, and using a linear \gls{SVM}.

The performance is shown using \gls{ROC} curves that indicate the performance of
binary classifiers at various false positive rates~\cite{Fawcett2006}.
Additionally, the plots show the \gls{AUC} scores for the top-4 features with
the highest \gls{AUC} emphasized in bold. In all three cases, KAZE/VGG
outperforms other feature combinations with an \gls{AUC} of~\SI{88.51}{\percent}
on all modules, followed by KAZE/SIFT with an \gls{AUC}
of~\SI{87.2175446838849}{\percent}. As an exception, the second best feature
combination for polycrystalline solar cells in terms of \gls{AUC} is \gls{PHOW}.
The gray dashed curve represents the baseline in terms of a random classifier.
Overall, the use of keypoints leads to better performance than dense sampling.

\subsection{Support Vector Machine vs.\ Transfer Learning Using Deep
Regression Network}

\Cref{fig:comparison_svm_cnn} shows the performance of the strongest \gls{SVM}
variant, KAZE/VGG, in comparison to the \gls{CNN} classifier. The \gls{ROC}
curve on the left in \cref{fig:roc-final-mono} contains the results for
monocrystalline \gls{PV} modules.  \Cref{fig:roc-final-poly} in the center
provides the classification performance for polycrystalline \gls{PV} modules.
Finally, the overall classification performance of both models is shown in
\cref{fig:roc-final-overall} on the right.

Notably, the classification performance of \gls{SVM} and \gls{CNN} is very
similar for monocrystalline \gls{PV} modules. The \gls{CNN} performs on average
only slightly better than the \gls{SVM}. At lower false positive rates around
and below~\SI{1}{\percent}, the \gls{CNN} achieves a higher true positive rate.
In the range of roughly~\SIrange{1}{10}{\percent} false positive rate, however,
the \gls{SVM} performs better.  This shows that KAZE/VGG is able to capture
surface abnormalities on homogeneous surfaces almost as accurate as a \gls{CNN}
trained on image pixels directly.

For polycrystalline \gls{PV} modules, the \gls{CNN} is able to predict
defective solar cells almost~\SI{11}{\percent} more accurately than the
\gls{SVM} in terms of the \gls{AUC}. This is also clearly a more difficult test
due to the large variety of textures among the solar cells.

Overall, the \gls{CNN} outperforms the \gls{SVM}. However, the performances of
both classifiers differ in total by only about \SI{6}{\percent}. The \gls{SVM}
classifier can therefore also be useful for a quick, on-the-spot assessment of a
\gls{PV} module in situations where specialized hardware for a \gls{CNN} is not
available.

\subsection{Model Performance per Defect Category}
\label{sec:confusion-matrices}

Here, we provide a detailed report of the performance of proposed models with
respect to individual categories of solar cells~(\ie, defective and functional)
in terms of confusion matrices. The two dimensional confusion matrix stores the
proportion of correctly identified cells (true negatives and true positives) in
each category on its primary diagonal. The secondary diagonal provides the
proportion of incorrectly identified solar cells (false negatives and false
positives) with respect to the other category.

\begin{figure*}[tbp]
  \centering
  \tikzsetnextfilename{cm-per-module-type}
  \includegraphics[width=\linewidth]{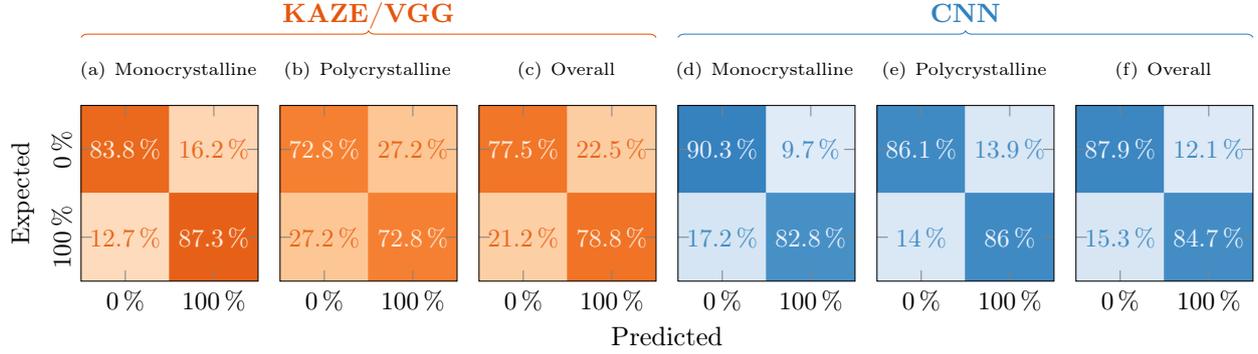}%
  \caption{Confusion matrices for the proposed classification models. Each row
    of confusion matrices stores the relative frequency of instances in the
    expected defect likelihood categories. The columns, on the other hand,
    contain the relative frequency of instances of predictions made by the
    classification models. Ideally, only the diagonals of confusion matrices
    would contain non-zero entries which corresponds to perfect agreement in all
    categories between the ground truth and the classification model. The
  \gls{CNN} generally makes less prediction errors than an \gls{SVM} trained on
KAZE/VGG features.}
  \label{fig:pv-cell-confusion-matrices}
\end{figure*}

\Cref{fig:pv-cell-confusion-matrices} shows the confusion matrices for the
proposed models. The confusion matrices are given for each type of
solar wafers, and their combination. The vertical axis of a confusion matrices
specifies the expected (\ie, ground truth) labels, whereas the horizontal one
the labels predicted by the corresponding model. Here, the predictions of the
\gls{CNN} were thresholded at \SI{50}{\percent} to produce the two categories of
functional~(\SI{0}{\percent}) and defective~(\SI{100}{\percent}) solar cells.

In regard to monocrystalline \gls{PV} modules, the confusion matrices in
\cref{fig:cm-lsvc-mono,fig:cm-cnn-mono} underline that both models provide
comparable classification results. The linear \gls{SVM}, however, is able to
identify more defective cells correctly than the \gls{CNN} at the expense of
functional cells being identified as defective (false negatives). To this end,
the linear \gls{SVM} makes also less errors at identifying defective solar cells
as being intact (false positives).

In polycrystalline case given by \cref{fig:cm-lsvc-poly,fig:cm-cnn-poly}, the
\gls{CNN} clearly outperforms the linear \gls{SVM} in every category. This also
leads to overall better performance of the \gls{CNN} in both cases, as evidenced
in \cref{fig:cm-lsvc-overall,fig:cm-cnn-overall}.

\subsection{Impact of Training Dataset Size on Model Performance}
\label{sec:subsets}

\begin{figure*}[tbp]
\centering
\tikzsetnextfilename{subset-scores}%
\includegraphics[width=\linewidth]{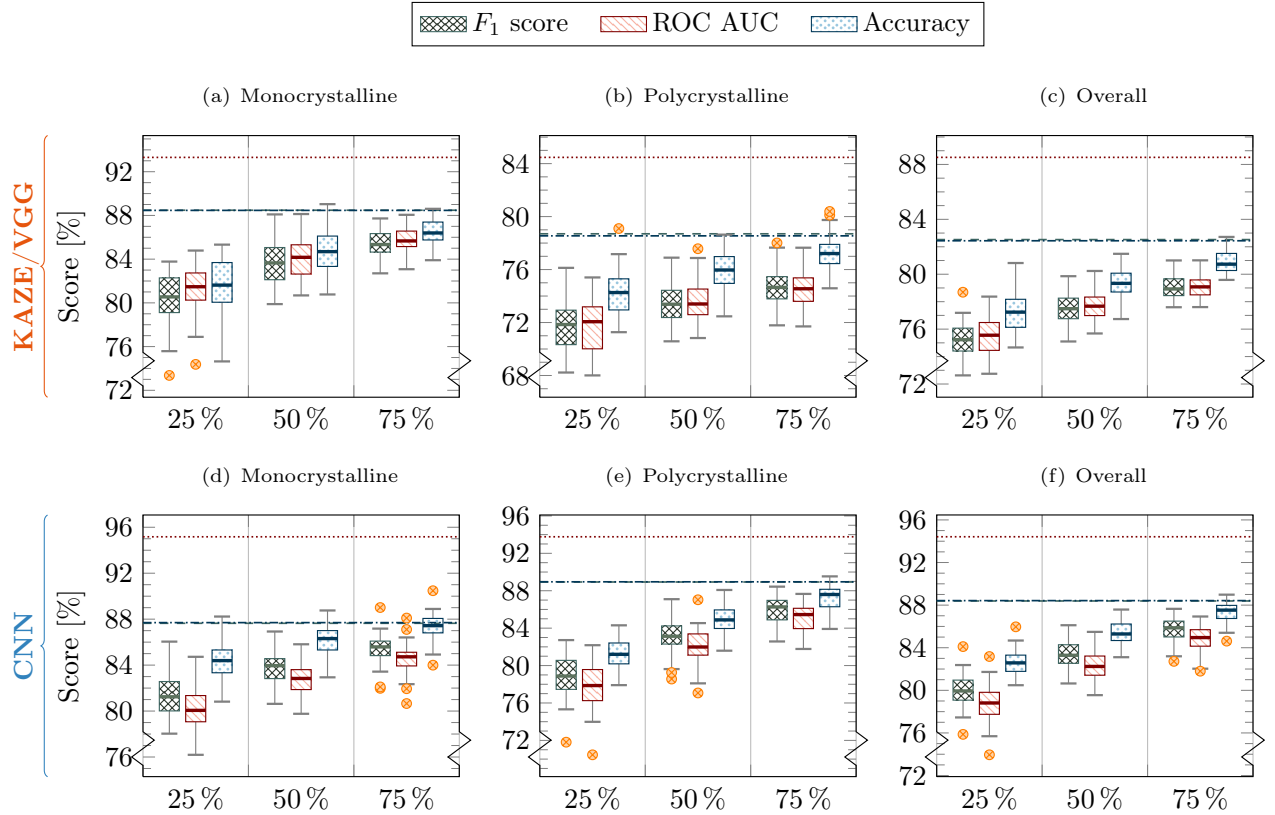}%
\caption{Performance of the proposed models trained on subsets of original
  training samples. The results are grouped by the solar wafer type (left two
  columns) and the combination of both wafer types (last column). The first
  three plots in the top row show the distribution of evaluated metrics as
  boxplots for the linear \gls{SVM} trained using KAZE/VGG features. The bottom
  row shows the results for the \gls{CNN}. The horizontal lines specify the
  reference scores with respect to the
  \(F_1\)~measure~(\ref*{pgfplots:f1-score-baseline}), \gls{ROC}
  \gls{AUC}~(\ref*{pgfplots:auc-score-baseline}), and the
  accuracy~(\ref*{pgfplots:accuracy-baseline}) of the proposed models trained on
  \SI{100}{\percent} of training samples. The circles~(\outliermarker) denote
  outliers in the distribution of evaluated metrics given by each boxplot.
  Increasing the number of training samples directly improves the performance of
  both models. The improvement is approximately logarithmic with respect to the
number of training samples.}
\label{fig:subset-scores}
\end{figure*}

For training both the linear \gls{SVM} and the \gls{CNN} a relatively small
dataset of unique solar cell images was used. Given that typical \gls{PV} module
production lines have an output of \num{1500} modules per day containing around
\num{90000} solar cells, models can be expected to greatly benefit from
additional training data. In order to examine how the proposed models improve if
more training samples are used, we evaluate their performance on subsets of
original training samples since no additional training samples are available.

To a infer the performance trend, we evaluate the models on three differently
sized subsets of original trainings samples. We used \SI{25}{\percent},
\SI{50}{\percent} and \SI{75}{\percent} of original training samples. To avoid a
bias in the obtained metrics, we not only sample the subsets randomly but also
sample each subset 50~times to obtain the samples used to train the models. We
additionally use stratified sampling to retain the distribution of labels from
the original set of training samples. To evaluate the performance, we use the
original test samples and provide the results for three metrics: \(F_1\)~score,
\gls{ROC} \gls{AUC}, and accuracy.

\Cref{fig:subset-scores} shows the distribution of evaluated scores on all
samples of the three differently sized subsets of training samples used to train
the proposed models. The distribution of all 50~scores for each of the three
subsets is summarized in a boxplot. The results clearly show that the
performance of the proposed models improves roughly logarithmically with respect
to the number of training samples which is typically observed in vision
tasks~\cite{Sun2017}.

\subsection{Analysis of the CNN Feature Space}

Here, we analyze the features learned by the \gls{CNN} using
\gls{t-SNE}~\cite{Maaten2008}, a manifold learning technique for dimensionality
reduction. The purpose is to examine the criteria for separation of different
solar cell clusters.  To this end, we use the Barnes-Hut
variant~\cite{Maaten2014} of \gls{t-SNE} which is substantially faster than the
standard implementation.  For computing the embeddings, we fixed \gls{t-SNE}'s
perplexity parameter to~35. Due to the small size of our test dataset, we
avoided an initial dimensionality reduction of the features using \gls{PCA} in
the preprocessing step, but rather used random initialization of embeddings.

\begin{figure}[tb]
  \centering
  \tikzsetnextfilename{tsne-hidden-layer}
  \includegraphics[width=\linewidth]{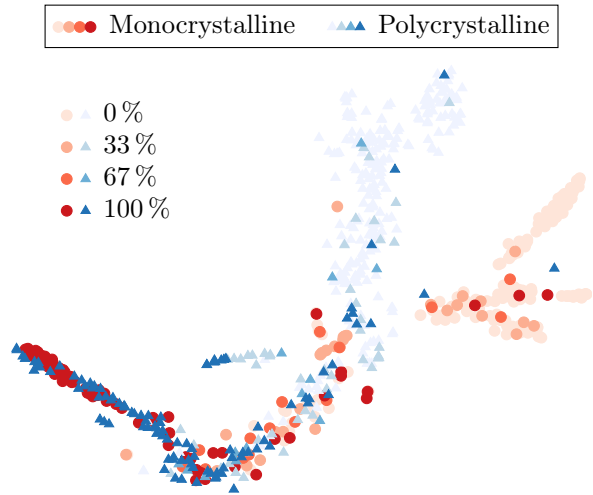}%
  \caption{\gls{t-SNE} visualization of the \gls{CNN}'s last hidden layer output
    for the four defect probability classes quantized from predictions of the
    deep regression network. The \num{2048}-dimensional output layer is mapped
    to a 2-D space for all \num{656}~test images. This structure preserving 2-D
  projection of embeddings shows that similar cells defects are grouped together
allowing the \gls{CNN} to discern between various defects.}
  \label{fig:pv-cell-defects-tsne}
\end{figure}

The resulting representation for all \num{656}~test images is shown in
\cref{fig:pv-cell-defects-tsne}. Each point corresponds to a feature vector
projected from the \num{2048}-dimensional last layer of the CNN onto two
dimensions.
\addition[label=a:tsne,ref=c:tsne]{Projected feature vectors that were extracted from mono-
and polycrystalline modules are color-coded in red and blue, respectively.
The defect probabilities are encoded by saturation.
The two dimensional
representation preserves the structure of the high-dimensional feature space
and shows that similar defect probabilities are in most cases co-located in
the features space.  This allows the \gls{CNN} classifier to distinguish
between \gls{EL} images of defective and functional solar cells.}

An important observation is that the class of definitely
defective~(\SI{100}{\percent}) cells forms a single elongated cluster (bottom
left) that includes cells irrespective of the source \gls{PV} module type. In
contrast to this, definitely functional cells~(\SI{0}{\percent}) are separated
into different clusters which depend on the type of the source \gls{PV} module.
The overall appearance of the cell (\ie, the number of soldering joints,
textureness, \etc) additionally generates several branches in the
monocrystalline cluster (on the right). These branches include samples grouped
by the number of busbar soldering joints within the cell. Here, the branches are
more pronounced than the separations in the cluster of functional
polycrystalline cells (at the top right) due to the homogeneous (\ie,
textureless) surface of the silicon wafers.

\begin{figure*}[tb]
  \centering
  \tikzset{
    module group/.style={
      draw=gray,
      rectangle,
      semithick,
      densely dashed,
      inner xsep=2ex,
      inner ysep=1ex,
    },
    mono module group/.style={module group},
    poly module group/.style={module group},
    group label/.style={
      gray,
      above,
      rotate=90,
    },
    correct incorrect/.style={
      draw,
      rectangle,
      inner sep=1ex,
    },
    correct/.style={
      correct incorrect,
      draw=gray!50,fill=gray!10,
    },
    incorrect/.style={
      correct incorrect,
      draw=OrangeRed3,fill=Tomato!30,
    },
    adjust near node color/.style={%
      defect score node=#1*10,
      font=\scriptsize,
      opacity=0.85,
    },
  }
  \pgfplotsset{
    colormap/Reds,
    plot gap/.style={xshift=2ex},
  }
  \tikzsetnextfilename{qualitative-results}%
  \includegraphics[width=\linewidth]{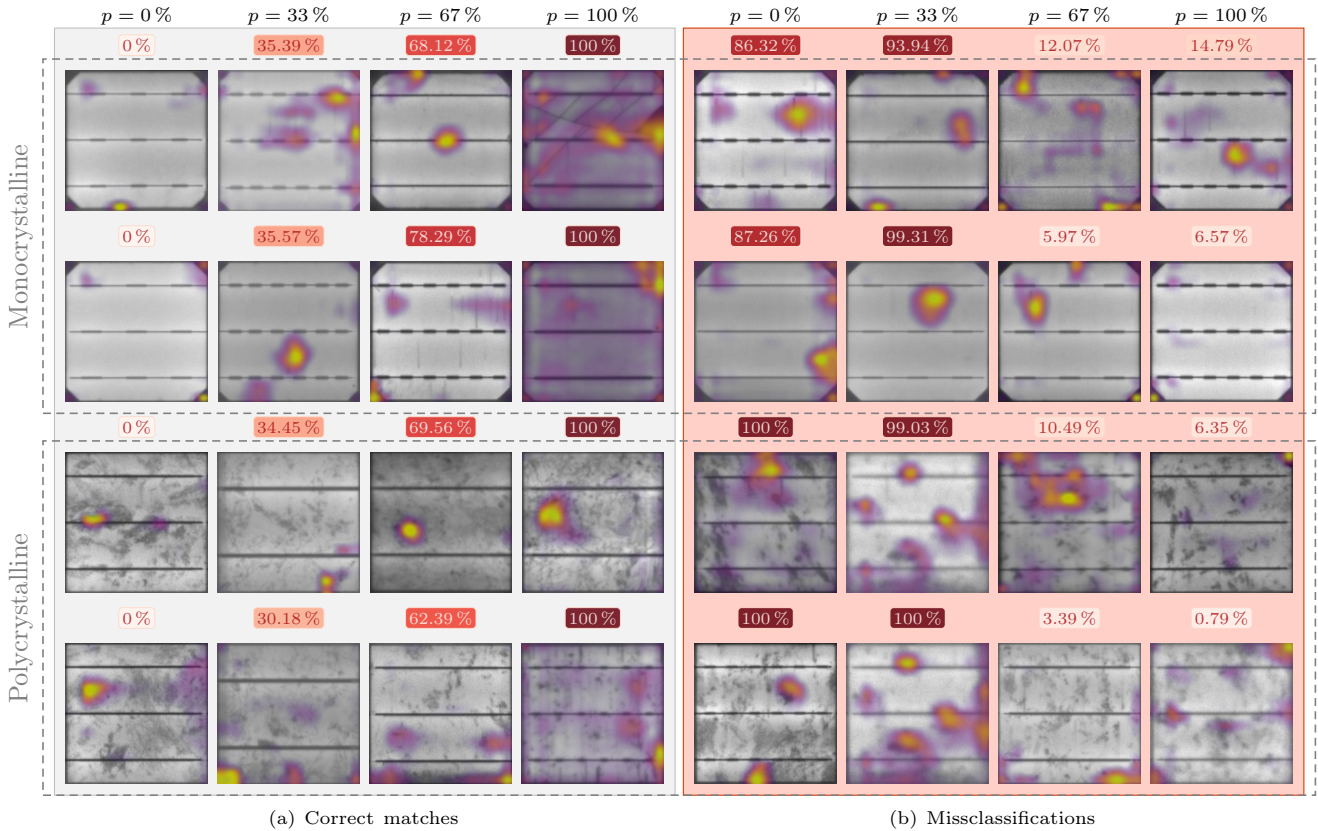}
  \caption{Qualitative results of predictions made by the proposed \gls{CNN}
    with correctly classified solar cell
    images~\subref{fig:correct-qualitative-results} and
    missclassifications~\subref{fig:incorrect-qualitative-results}. Each column
    is labeled using the ground truth label. Red shaded probabilities above each
    solar cell image correspond to predictions made by the \gls{CNN}. The upper
    two rows correspond to monocrystalline solar cells and bottom two rows to
  polycrystalline solar cell images.}
  \label{fig:qualitative-results}
\end{figure*}

The clusters for the categories of possibly defective~(\SI{33}{\percent}) and
likely defective~(\SI{67}{\percent}) cells are mixed. The high confusion
between these samples stems from the comparably small size of the corresponding
categories compared to the size of the two remaining categories of high
confidence samples in our dataset (see \cref{tab:pv-cell-labels-distribution}).
Additionally, the samples from these two categories can stimulate ambiguous
decisions due to being at the boundary of clearly distinguishable defects and
non-defects.

\subsection{Qualitative Results}
\label{sec:qualitative-results}

\Cref{fig:qualitative-results} provides qualitative results for a selection of
monocrystalline and polycrystalline solar cells with the corresponding defect
likelihoods inferred by the proposed \gls{CNN}. To allow an easy comparison to
ground truth labels, the \gls{CNN} defect probabilities are quantized into four
categories corresponding to original labels by rounding the probabilities to
nearest category. The selection contains both correctly as well as incorrectly
classified solar cells having the smallest and largest squared distance,
respectively, between the predicted probability and the ground truth label.

In order to highlight class-specific discriminative regions in solar cell
images, \glspl{CAM}~\cite{Zhou2016,Selvaraju2017,Chattopadhay2018} can be
employed. While \glspl{CAM} are not directly suitable for precise segmentation
of defective regions particularly due to their coarse resolution. \Glspl{CAM}
can still provide cues that explain why the convolutional network infers a
specific defect probability. To this end, the solar cells in
\cref{fig:qualitative-results} are additionally overlaid by \glspl{CAM}
extracted from the last convolutional block~(\(18\times18\times512\)) of the
modified \vggnet\ network and upscaled to the original resolution
of~\(300\times300\) of solar cell images using the methodology
by~\citet{Chattopadhay2018}.

\begin{figure*}[tb]
  \tikzsetnextfilename{annotated-qualitative-example}
  \centering
  \includegraphics[width=\linewidth]{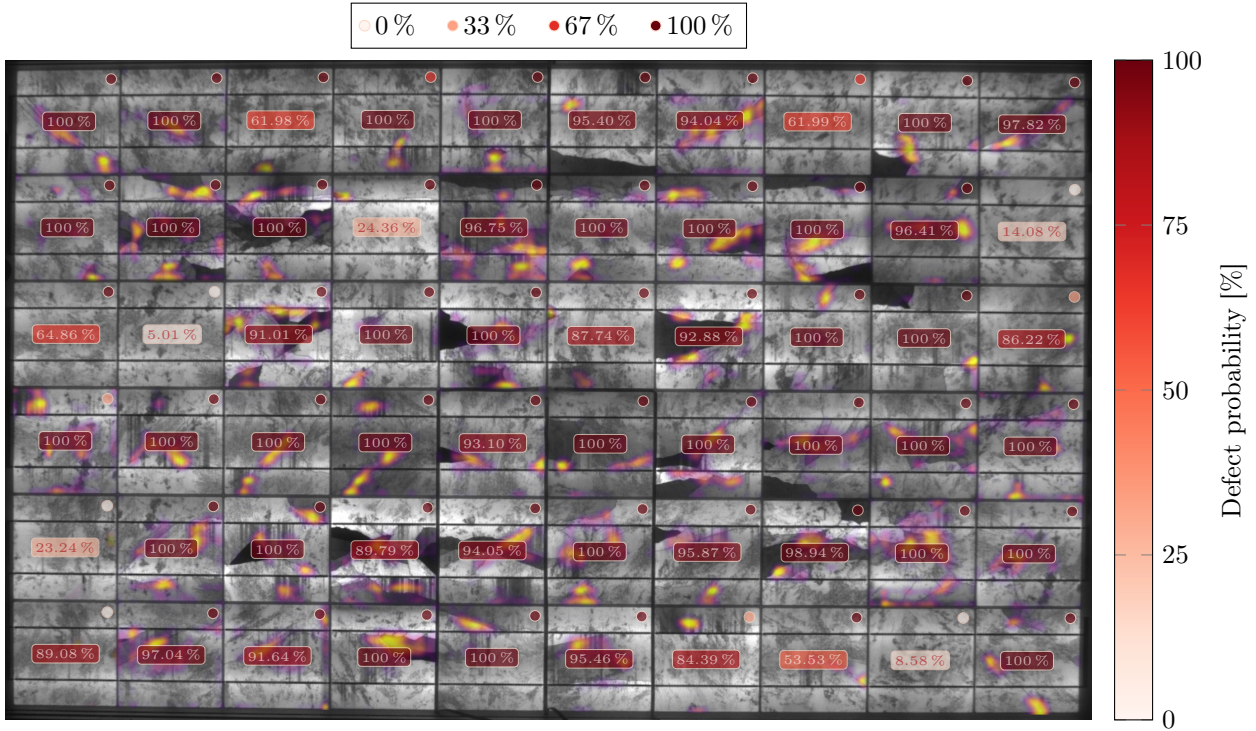}
  \caption{Qualitative defect classification results in a \gls{PV} module
    previously not seen by the deep regression network. The red shaded circles
    in the top right corner of each solar cell specify the ground truth labels.
    The solar cells are additionally overlaid by \glspl{CAM} determined using
    Grad-CAM++~\cite{Chattopadhay2018}. The \gls{CAM} for individual solar cells
    was additionally weighted by network's predictions to reduce the clutter.
    Notably, the network pays attention to very specific defects (such as fine
  cell cracks) that are harder to identify than cell cracks which are more
obvious.}
  \label{fig:qualtiative-pv-module}
\end{figure*}

Interestingly, even if the \gls{CNN} incorrectly classifies a defective solar
cell to be functional (\cf, last column in
\cref{fig:incorrect-qualitative-results}), the \gls{CAM} can still highlight
image regions which are potentially defective. \Glspl{CAM} can therefore
complement the fully automatic assessment process and provide decision support
in complicated situations during visual inspection. One particular problem that
can be witnessed from inspection of \glspl{CAM} is that finger interruptions are
not always clearly discerned from actual defects. This, however, can be managed
by including corresponding samples to train the \gls{CNN}.

In \cref{fig:qualtiative-pv-module} we show the predictions of the \gls{CNN} for
a complete polycrystalline solar module. The ground truth labels are given as
red shaded circles in the top right corner of each solar cell. Again, the solar
cells are overlaid by \glspl{CAM} and additionally weighted by network's
predictions to reduce the amount of visual clutter. By inspecting the
\glspl{CAM} it can be observed that the \gls{CNN} focuses on particularly unique
defects within solar cells that are harder to identify than more obvious defects
such as degraded or electrically insulated cell parts (appearing as dark
regions) in the same cell.

\subsection{Runtime Evaluation}
\label{sec:runtime-evaluation}

\begin{figure}[tb]
  \centering
  \tikzsetnextfilename{timings-overview}%
  \includegraphics[width=\linewidth]{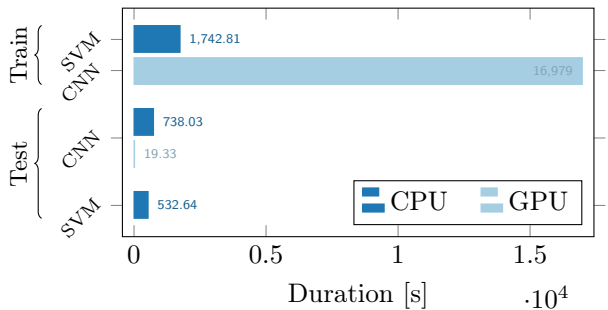}
  \caption{Runtime of training and test phases for the proposed models. Training
    the \gls{SVM} takes around 30~minutes, whereas training the \gls{CNN} takes
    almost 5~hours. The \gls{CNN} is overall more efficient at inference (\ie,
    testing) when running on the \gls{GPU} requiring just slightly less than
    20~seconds compared to over 12~minutes on the CPU. Our unoptimized
  implementation of the \gls{SVM} pipeline completes in only 8~minutes.}
  \label{fig:timings}
\end{figure}

\begin{figure}[tb]
  \centering
  \tikzsetnextfilename{timings-svm}%
  \includegraphics[width=\linewidth]{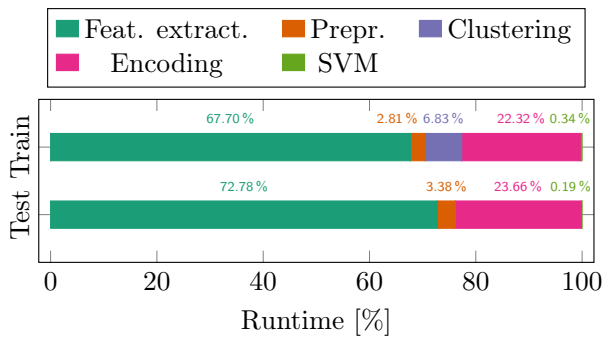}
  \caption{Relative runtime contributions to training and test phases of the
  \gls{SVM} pipeline. The most time-demanding step during \gls{SVM} training and
inference is the detection and extraction of KAZE/VGG features.}
  \label{fig:timings-svm}
\end{figure}

Here, we evaluate the time taken by each step of the \gls{SVM} pipeline and by
the \gls{CNN}, both during training and testing. The runtime is evaluated on a
system running an Intel i7-3770K CPU clocked at~\SI{3.50}{\giga\hertz} with
\SI{32}{\giga\byte} of RAM. The results are summarized in \cref{fig:timings}.

Unsurprisingly, training takes most of the time for both models. While training
the \gls{SVM} takes in total around 30~minutes. Refining the \gls{CNN} is almost
ten times slower and takes around 5~hours. However, inference using \gls{CNN} is
much faster than that of the \gls{SVM} pipeline and takes just under 20~seconds
over 8~minutes of the \gls{SVM}. It is, however, important to note that the
\gls{SVM} pipeline inference duration is reported for the execution on the CPU,
whereas the duration of the much faster \gls{CNN} inference is obtained on the
\gls{GPU} only. Additionally, only a part of the \gls{SVM} pipeline performs the
processing in parallel. When running the highly parallel \gls{CNN} inference on
the CPU, the test time increases considerably to over 12~minutes. Consequently,
training the \gls{CNN} on the CPU becomes intractable and we therefore refrained
from measuring the corresponding runtime.

Considering the relative contributions of individual \gls{SVM} pipeline steps,
feature extraction is most time-consuming, followed by encoding of local
features and clustering (\cf, \cref{fig:timings-svm}). Preprocessing of features
and hyperparameter optimization require the least.

In applications that require not only a low resource footprint but also must run
fast, the total execution time of the \gls{SVM} pipeline can be reduced by
replacing the VGG feature descriptor either by \gls{SIFT} or \gls{PHOW}. Both
feature descriptors substantially reduce the time taken for feature extraction
during inference from originally 8~minutes to around 23~seconds and 12~seconds,
respectively while maintaining a classification performance similar to the VGG
descriptor.

\subsection{Discussion}

Several conclusions can be drawn from the evaluation results. First, masking can
be useful if the spatial distribution of keypoints is rather sparse. However, in
most cases masking does not improve the classification accuracy. Secondly,
weighting samples proportionally to the confidence of the defect likelihood in a
cell does improve the generalization ability of the learned classifiers.

KAZE/VGG features trained using linear \gls{SVM} is the best performing
\gls{SVM} pipeline variant with an accuracy of~\SI{82.43847875}{\percent} and an
\(F_1\)~score of~\SI{82.52}{\percent}.  The \gls{CNN} is even more accurate. It
distinguishes functional and defective solar cells with an accuracy
of~\SI{88.4228187919}{\percent}. The corresponding \(F_1\)~score
is~\SI{88.3898929859}{\percent}. The 2-dimensional visualization of the
\gls{CNN} feature distribution via \gls{t-SNE} underlines that the network
learns the actual structure of the task at hand.

A limitation of the proposed method is that each solar cell is examined
independently. In particular, some types of surface abnormalities that do not
affect the module efficiency can appear in repetitive patterns across cells.
Accurate classification of such larger-scale effects requires to take context
into consideration, which is subject to future work.

\addition[label=a:defect-types,ref=c:defect-types]{Instead of predicting the
  defect likelihood one may want to predict specific defect types. Given
  additional training data, the methodology presented in this work can be
  applied without major changes (\eg, by fine-tuning to the new defect
  categories) given additional training data with appropriate labels.
  Fine-tuning the network to multiple defect categories with the goal of
  predicting defect types instead of their probabilities, however, will
  generally affect the choice of the loss function and consequently the number
of neurons in the last activation layer. A common choice for the loss function for
such tasks is the (categorical) cross entropy loss with softmax
activation~\cite{Goodfellow2016}.}

\section{Conclusions}
\label{sec:pv-cell-conclusions}

We presented a general framework for training an \gls{SVM} and a \gls{CNN} that
can be employed for identifying defective solar cells in high resolution
\gls{EL} images. The processing pipeline for the \gls{SVM} classifier is
carefully designed. In a series of experiments, the best performing pipeline is
determined as KAZE/VGG features in a linear \gls{SVM} trained on samples that
take the confidence of the labeler into consideration. The \gls{CNN} network is
a fine-tuned regression network based on \vggnet, trained on augmented cell images
that also consider the labeler confidence.

On monocrystalline solar modules, both classifiers perform similarly well, with
only a slight advantage on average for the \gls{CNN}.  However, the \gls{CNN}
classifier outperforms the \gls{SVM} classifier by about \SI{6}{\percent}
accuracy on the more inhomogeneous polycrystalline cells. This leads also to
the better average accuracy across all cells of~\SI{88.4228187919}{\percent}
for the \gls{CNN} versus \SI{82.43847875}{\percent} for the \gls{SVM}. The high
accuracies make both classifiers useful for visual inspection. If the
application scenario permits the usage of \glspl{GPU} and higher processing
times, the computationally more expensive \gls{CNN} is preferred.  Otherwise,
the \gls{SVM} classifier is a viable alternative for applications that require
a low resource footprint.

\section*{Acknowledgments}

This work was funded by Energie Campus Nürnberg~(EnCN) and partially supported
by the Research Training Group~1773 \enquote{Heterogeneous Image Systems} funded
by the German Research Foundation~(DFG).


\bibliographystyle{elsarticle-num-names}
\let\tt\ttfamily
\bibliography{references}

\begin{thebibliography}{62}
\expandafter\ifx\csname natexlab\endcsname\relax\def\natexlab#1{#1}\fi
\providecommand{\url}[1]{\texttt{#1}}
\providecommand{\href}[2]{#2}
\providecommand{\path}[1]{#1}
\providecommand{\DOIprefix}{doi:}
\providecommand{\ArXivprefix}{arXiv:}
\providecommand{\URLprefix}{URL: }
\providecommand{\Pubmedprefix}{pmid:}
\providecommand{\doi}[1]{\href{http://dx.doi.org/#1}{\path{#1}}}
\providecommand{\Pubmed}[1]{\href{pmid:#1}{\path{#1}}}
\providecommand{\bibinfo}[2]{#2}
\ifx\xfnm\relax \def\xfnm[#1]{\unskip,\space#1}\fi
\bibitem[{Kajari-Schr{{{{\"{o}}}}}der et~al.(2012)Kajari-Schr{{{{\"{o}}}}}der,
  Kunze, and K{{{{\"{o}}}}}ntges}]{Kajari-Schroder2012}
\bibinfo{author}{S.~Kajari-Schr{{{{\"{o}}}}}der}, \bibinfo{author}{I.~Kunze},
  \bibinfo{author}{M.~K{{{{\"{o}}}}}ntges},
\newblock \bibinfo{title}{Criticality of cracks in {PV} modules},
\newblock \bibinfo{journal}{Energy Procedia} \bibinfo{volume}{27}
  (\bibinfo{year}{2012}) \bibinfo{pages}{658--663}.
  \DOIprefix\doi{10.1016/j.egypro.2012.07.125}.
\bibitem[{Fuyuki et~al.(2005)Fuyuki, Kondo, Yamazaki, Takahashi, and
  Uraoka}]{Fuyuki2005}
\bibinfo{author}{T.~Fuyuki}, \bibinfo{author}{H.~Kondo},
  \bibinfo{author}{T.~Yamazaki}, \bibinfo{author}{Y.~Takahashi},
  \bibinfo{author}{Y.~Uraoka},
\newblock \bibinfo{title}{Photographic surveying of minority carrier diffusion
  length in polycrystalline silicon solar cells by electroluminescence},
\newblock \bibinfo{journal}{Applied Physics Letters} \bibinfo{volume}{86}
  (\bibinfo{year}{2005}). \DOIprefix\doi{10.1063/1.1978979}.
\bibitem[{Fuyuki and Kitiyanan(2009)}]{Fuyuki2009}
\bibinfo{author}{T.~Fuyuki}, \bibinfo{author}{A.~Kitiyanan},
\newblock \bibinfo{title}{Photographic diagnosis of crystalline silicon solar
  cells utilizing electroluminescence},
\newblock \bibinfo{journal}{Applied Physics A} \bibinfo{volume}{96}
  (\bibinfo{year}{2009}) \bibinfo{pages}{189--196}.
  \DOIprefix\doi{10.1007/s00339-008-4986-0}.
\bibitem[{Breitenstein et~al.(2011)Breitenstein, Bauer, Bothe, Hinken,
  M{{{{\"{u}}}}}ller, Kwapil, Schubert, and Warta}]{Breitenstein2011}
\bibinfo{author}{O.~Breitenstein}, \bibinfo{author}{J.~Bauer},
  \bibinfo{author}{K.~Bothe}, \bibinfo{author}{D.~Hinken},
  \bibinfo{author}{J.~M{{{{\"{u}}}}}ller}, \bibinfo{author}{W.~Kwapil},
  \bibinfo{author}{M.~C. Schubert}, \bibinfo{author}{W.~Warta},
\newblock \bibinfo{title}{Can luminescence imaging replace lock-in thermography
  on solar cells?},
\newblock \bibinfo{journal}{IEEE Journal of Photovoltaics} \bibinfo{volume}{1}
  (\bibinfo{year}{2011}) \bibinfo{pages}{159--167}.
  \DOIprefix\doi{10.1109/JPHOTOV.2011.2169394}.
\bibitem[{K{{{{\"{o}}}}}ntges et~al.(2014)K{{{{\"{o}}}}}ntges, Kurtz, Packard,
  Jahn, Berger, Kato, Friesen, Liu, and {Van Iseghem}}]{Koentges2014}
\bibinfo{author}{M.~K{{{{\"{o}}}}}ntges}, \bibinfo{author}{S.~Kurtz},
  \bibinfo{author}{C.~Packard}, \bibinfo{author}{U.~Jahn},
  \bibinfo{author}{K.~Berger}, \bibinfo{author}{K.~Kato},
  \bibinfo{author}{T.~Friesen}, \bibinfo{author}{H.~Liu},
  \bibinfo{author}{M.~{Van Iseghem}}, \bibinfo{title}{Review of Failures of
  Photovoltaic Modules}, \bibinfo{type}{Technical Report},
  \bibinfo{year}{2014}.
\bibitem[{{De Rose} et~al.(2012){De Rose}, Malomo, Magnone, Crupi, Cellere,
  Martire, Tonini, and Sangiorgi}]{DeRose2012}
\bibinfo{author}{R.~{De Rose}}, \bibinfo{author}{A.~Malomo},
  \bibinfo{author}{P.~Magnone}, \bibinfo{author}{F.~Crupi},
  \bibinfo{author}{G.~Cellere}, \bibinfo{author}{M.~Martire},
  \bibinfo{author}{D.~Tonini}, \bibinfo{author}{E.~Sangiorgi},
\newblock \bibinfo{title}{A methodology to account for the finger interruptions
  in solar cell performance},
\newblock \bibinfo{journal}{Microelectronics Reliability} \bibinfo{volume}{52}
  (\bibinfo{year}{2012}) \bibinfo{pages}{2500--2503}.
  \DOIprefix\doi{10.1016/j.microrel.2012.07.014}.
\bibitem[{Kang and Cha(2018)}]{Kang2018}
\bibinfo{author}{D.~Kang}, \bibinfo{author}{Y.-J. Cha},
\newblock \bibinfo{title}{Autonomous {UAVs} for structural health monitoring
  using deep learning and an ultrasonic beacon system with geo-tagging},
\newblock \bibinfo{journal}{Computer-Aided Civil and Infrastructure
  Engineering} \bibinfo{volume}{33} (\bibinfo{year}{2018})
  \bibinfo{pages}{885--902}. \DOIprefix\doi{10.1111/mice.12375}.
\bibitem[{Dotenco et~al.(2016)Dotenco, Dalsass, Winkler, W{\"{u}}rzner, Brabec,
  Maier, and Gallwitz}]{Dotenco2016}
\bibinfo{author}{S.~Dotenco}, \bibinfo{author}{M.~Dalsass},
  \bibinfo{author}{L.~Winkler}, \bibinfo{author}{T.~W{\"{u}}rzner},
  \bibinfo{author}{C.~Brabec}, \bibinfo{author}{A.~Maier},
  \bibinfo{author}{F.~Gallwitz},
\newblock \bibinfo{title}{Automatic detection and analysis of photovoltaic
  modules in aerial infrared imagery},
\newblock in: \bibinfo{booktitle}{Winter Conference on Applications of Computer
  Vision (WACV)}, \bibinfo{year}{2016}, p.~\bibinfo{pages}{9}.
  \DOIprefix\doi{10.1109/WACV.2016.7477658}.
\bibitem[{Tsai et~al.(2012)Tsai, Wu, and Li}]{Tsai2012}
\bibinfo{author}{D.-M. Tsai}, \bibinfo{author}{S.-C. Wu},
  \bibinfo{author}{W.-C. Li},
\newblock \bibinfo{title}{Defect detection of solar cells in
  electroluminescence images using {Fourier} image reconstruction},
\newblock \bibinfo{journal}{Solar Energy Materials and Solar Cells}
  \bibinfo{volume}{99} (\bibinfo{year}{2012}) \bibinfo{pages}{250--262}.
  \DOIprefix\doi{10.1016/j.solmat.2011.12.007}.
\bibitem[{Tsai et~al.(2013)Tsai, Wu, and Chiu}]{Tsai2013}
\bibinfo{author}{D.-M. Tsai}, \bibinfo{author}{S.-C. Wu},
  \bibinfo{author}{W.-Y. Chiu},
\newblock \bibinfo{title}{Defect detection in solar modules using {ICA} basis
  images},
\newblock \bibinfo{journal}{IEEE Transactions on Industrial Informatics}
  \bibinfo{volume}{9} (\bibinfo{year}{2013}) \bibinfo{pages}{122--131}.
  \DOIprefix\doi{10.1109/TII.2012.2209663}.
\bibitem[{Anwar and Abdullah(2014)}]{Anwar2014}
\bibinfo{author}{S.~A. Anwar}, \bibinfo{author}{M.~Z. Abdullah},
\newblock \bibinfo{title}{Micro-crack detection of multicrystalline solar cells
  featuring an improved anisotropic diffusion filter and image segmentation
  technique},
\newblock \bibinfo{journal}{EURASIP Journal on Image and Video Processing}
  \bibinfo{volume}{2014} (\bibinfo{year}{2014}) \bibinfo{pages}{15}.
  \DOIprefix\doi{10.1186/1687-5281-2014-15}.
\bibitem[{Tseng et~al.(2015)Tseng, Liu, and Chou}]{Tseng2015}
\bibinfo{author}{D.-C. Tseng}, \bibinfo{author}{Y.-S. Liu},
  \bibinfo{author}{C.-M. Chou},
\newblock \bibinfo{title}{Automatic finger interruption detection in
  electroluminescence images of multicrystalline solar cells},
\newblock \bibinfo{journal}{Mathematical Problems in Engineering}
  \bibinfo{volume}{2015} (\bibinfo{year}{2015}) \bibinfo{pages}{1--12}.
  \DOIprefix\doi{10.1155/2015/879675}.
\bibitem[{Mehta et~al.(2018)Mehta, Azad, Chemmengath, Raykar, and
  Kalyanaraman}]{Mehta2017}
\bibinfo{author}{S.~Mehta}, \bibinfo{author}{A.~P. Azad},
  \bibinfo{author}{S.~A. Chemmengath}, \bibinfo{author}{V.~Raykar},
  \bibinfo{author}{S.~Kalyanaraman},
\newblock \bibinfo{title}{{DeepSolarEye}: Power loss prediction and weakly
  supervised soiling localization via fully convolutional networks for solar
  panels},
\newblock in: \bibinfo{booktitle}{Winter Conference on Applications of Computer
  Vision (WACV)}, \bibinfo{year}{2018}, pp. \bibinfo{pages}{333--342}.
  \DOIprefix\doi{10.1109/WACV.2018.00043}.
\bibitem[{Masci et~al.(2012)Masci, Meier, Ciresan, Schmidhuber, and
  Fricout}]{Masci2012}
\bibinfo{author}{J.~Masci}, \bibinfo{author}{U.~Meier},
  \bibinfo{author}{D.~Ciresan}, \bibinfo{author}{J.~Schmidhuber},
  \bibinfo{author}{G.~Fricout},
\newblock \bibinfo{title}{Steel defect classification with max-pooling
  convolutional neural networks},
\newblock in: \bibinfo{booktitle}{International Joint Conference on Neural
  Networks (IJCNN)}, \bibinfo{year}{2012}, pp. \bibinfo{pages}{1--6}.
  \DOIprefix\doi{10.1109/IJCNN.2012.6252468}.
\bibitem[{Zhang et~al.(2016)Zhang, Yang, {Daniel Zhang}, and Zhu}]{Zhang2016}
\bibinfo{author}{L.~Zhang}, \bibinfo{author}{F.~Yang},
  \bibinfo{author}{Y.~{Daniel Zhang}}, \bibinfo{author}{Y.~J. Zhu},
\newblock \bibinfo{title}{Road crack detection using deep convolutional neural
  network},
\newblock in: \bibinfo{booktitle}{International Conference on Image Processing
  (ICIP)}, \bibinfo{year}{2016}, pp. \bibinfo{pages}{3708--3712}.
  \DOIprefix\doi{10.1109/ICIP.2016.7533052}.
\bibitem[{Cha et~al.(2017)Cha, Choi, and
  B{\"{u}}y{\"{u}}k{\"{o}}zt{\"{u}}rk}]{Cha2017}
\bibinfo{author}{Y.-J. Cha}, \bibinfo{author}{W.~Choi},
  \bibinfo{author}{O.~B{\"{u}}y{\"{u}}k{\"{o}}zt{\"{u}}rk},
\newblock \bibinfo{title}{Deep learning-based crack damage detection using
  convolutional neural networks},
\newblock \bibinfo{journal}{Computer-Aided Civil and Infrastructure
  Engineering} \bibinfo{volume}{32} (\bibinfo{year}{2017})
  \bibinfo{pages}{361--378}. \DOIprefix\doi{10.1111/mice.12263}.
\bibitem[{Cha et~al.(2018)Cha, Choi, Suh, Mahmoudkhani, and
  B{\"{u}}y{\"{u}}k{\"{o}}zt{\"{u}}rk}]{Cha2018}
\bibinfo{author}{Y.-J. Cha}, \bibinfo{author}{W.~Choi},
  \bibinfo{author}{G.~Suh}, \bibinfo{author}{S.~Mahmoudkhani},
  \bibinfo{author}{O.~B{\"{u}}y{\"{u}}k{\"{o}}zt{\"{u}}rk},
\newblock \bibinfo{title}{Autonomous structural visual inspection using
  region-based deep learning for detecting multiple damage types},
\newblock \bibinfo{journal}{Computer-Aided Civil and Infrastructure
  Engineering} \bibinfo{volume}{33} (\bibinfo{year}{2018})
  \bibinfo{pages}{731--747}. \DOIprefix\doi{10.1111/mice.12334}.
\bibitem[{Lee et~al.(2019)Lee, Kim, and Lee}]{Lee2019}
\bibinfo{author}{D.~Lee}, \bibinfo{author}{J.~Kim}, \bibinfo{author}{D.~Lee},
\newblock \bibinfo{title}{Robust concrete crack detection using deep
  learning-based semantic segmentation},
\newblock \bibinfo{journal}{International Journal of Aeronautical and Space
  Sciences}  (\bibinfo{year}{2019}). \DOIprefix\doi{10.1007/s42405-018-0120-5}.
\bibitem[{Esteva et~al.(2017)Esteva, Kuprel, Novoa, Ko, Swetter, Blau, and
  Thrun}]{Esteva2017}
\bibinfo{author}{A.~Esteva}, \bibinfo{author}{B.~Kuprel},
  \bibinfo{author}{R.~A. Novoa}, \bibinfo{author}{J.~Ko},
  \bibinfo{author}{S.~M. Swetter}, \bibinfo{author}{H.~M. Blau},
  \bibinfo{author}{S.~Thrun},
\newblock \bibinfo{title}{Dermatologist-level classification of skin cancer
  with deep neural networks},
\newblock \bibinfo{journal}{Nature} \bibinfo{volume}{542}
  (\bibinfo{year}{2017}) \bibinfo{pages}{115--118}.
  \DOIprefix\doi{10.1038/nature21056}.
\bibitem[{Deitsch et~al.(2018)Deitsch, Buerhop-Lutz, Maier, Gallwitz, and
  Riess}]{Deitsch2018}
\bibinfo{author}{S.~Deitsch}, \bibinfo{author}{C.~Buerhop-Lutz},
  \bibinfo{author}{A.~Maier}, \bibinfo{author}{F.~Gallwitz},
  \bibinfo{author}{C.~Riess}, \bibinfo{title}{Segmentation of Photovoltaic
  Module Cells in Electroluminescence Images}, \bibinfo{type}{e-print},
  \bibinfo{year}{2018}. \href{http://arxiv.org/abs/1806.06530}{{\tt
  arXiv:1806.06530}}.
\bibitem[{Cortes and Vapnik(1995)}]{Cortes1995}
\bibinfo{author}{C.~Cortes}, \bibinfo{author}{V.~Vapnik},
\newblock \bibinfo{title}{Support-vector networks},
\newblock \bibinfo{journal}{Machine Learning} \bibinfo{volume}{20}
  (\bibinfo{year}{1995}) \bibinfo{pages}{273--297}.
  \DOIprefix\doi{10.1007/BF00994018}.
\bibitem[{Mair et~al.(2010)Mair, Hager, Burschka, Suppa, and
  Hirzinger}]{Mair2010}
\bibinfo{author}{E.~Mair}, \bibinfo{author}{G.~D. Hager},
  \bibinfo{author}{D.~Burschka}, \bibinfo{author}{M.~Suppa},
  \bibinfo{author}{G.~Hirzinger},
\newblock \bibinfo{title}{Adaptive and generic corner detection based on the
  accelerated segment test},
\newblock in: \bibinfo{booktitle}{European Conference on Computer Vision
  (ECCV)}, volume \bibinfo{volume}{6312} of \textit{\bibinfo{series}{Lecture
  Notes in Computer Science}}, \bibinfo{year}{2010}, pp.
  \bibinfo{pages}{183--196}. \DOIprefix\doi{10.1007/978-3-642-15552-9_14}.
\bibitem[{Alcantarilla et~al.(2012)Alcantarilla, Bartoli, and
  Davison}]{Alcantarilla2012}
\bibinfo{author}{P.~F. Alcantarilla}, \bibinfo{author}{A.~Bartoli},
  \bibinfo{author}{A.~J. Davison},
\newblock \bibinfo{title}{{KAZE} features},
\newblock in: \bibinfo{booktitle}{European Conference on Computer Vision
  ({ECCV})}, volume \bibinfo{volume}{7577} of \textit{\bibinfo{series}{Lecture
  Notes in Computer Science}}, \bibinfo{year}{2012}, pp.
  \bibinfo{pages}{214--227}. \DOIprefix\doi{10.1007/978-3-642-33783-3_16}.
\bibitem[{Dalal and Triggs(2005)}]{Dalal2005}
\bibinfo{author}{N.~Dalal}, \bibinfo{author}{B.~Triggs},
\newblock \bibinfo{title}{Histograms of oriented gradients for human
  detection},
\newblock in: \bibinfo{booktitle}{Conference on Computer Vision and Pattern
  Recognition (CVPR)}, volume~\bibinfo{volume}{1}, \bibinfo{year}{2005}, pp.
  \bibinfo{pages}{886--893}. \DOIprefix\doi{10.1109/CVPR.2005.177}.
\bibitem[{Bosch et~al.(2007)Bosch, Zisserman, and Munoz}]{Bosch2007}
\bibinfo{author}{A.~Bosch}, \bibinfo{author}{A.~Zisserman},
  \bibinfo{author}{X.~Munoz},
\newblock \bibinfo{title}{Image classification using random forests and ferns},
\newblock in: \bibinfo{booktitle}{International Conference on Computer Vision
  (ICCV)}, \bibinfo{year}{2007}, pp. \bibinfo{pages}{1--8}.
  \DOIprefix\doi{10.1109/ICCV.2007.4409066}.
\bibitem[{Lowe(1999)}]{Lowe1999}
\bibinfo{author}{D.~G. Lowe},
\newblock \bibinfo{title}{Object recognition from local scale-invariant
  features},
\newblock in: \bibinfo{booktitle}{International Conference on Computer Vision
  (ICCV)}, volume~\bibinfo{volume}{2}, \bibinfo{year}{1999}, pp.
  \bibinfo{pages}{1150--1157}. \DOIprefix\doi{10.1109/ICCV.1999.790410}.
\bibitem[{Bay et~al.(2008)Bay, Essa, Tuytelaarsb, and Van~Goola}]{Bay2008}
\bibinfo{author}{H.~Bay}, \bibinfo{author}{A.~Essa},
  \bibinfo{author}{T.~Tuytelaarsb}, \bibinfo{author}{L.~Van~Goola},
\newblock \bibinfo{title}{Speeded-up robust features ({SURF})},
\newblock \bibinfo{journal}{Computer Vision and Image Understanding}
  \bibinfo{volume}{110} (\bibinfo{year}{2008}) \bibinfo{pages}{346--359}.
  \DOIprefix\doi{10.1016/j.cviu.2007.09.014}.
\bibitem[{Simonyan et~al.(2014)Simonyan, Vedaldi, and
  Zisserman}]{Simonyan2014a}
\bibinfo{author}{K.~Simonyan}, \bibinfo{author}{A.~Vedaldi},
  \bibinfo{author}{A.~Zisserman},
\newblock \bibinfo{title}{Learning local feature descriptors using convex
  optimisation},
\newblock \bibinfo{journal}{IEEE Transactions on Pattern Analysis and Machine
  Intelligence} \bibinfo{volume}{36} (\bibinfo{year}{2014})
  \bibinfo{pages}{1573--1585}. \DOIprefix\doi{10.1109/TPAMI.2014.2301163}.
\bibitem[{Rosten and Drummond(2005)}]{Rosten2005}
\bibinfo{author}{E.~Rosten}, \bibinfo{author}{T.~Drummond},
\newblock \bibinfo{title}{Fusing points and lines for high performance
  tracking},
\newblock in: \bibinfo{booktitle}{International Conference on Computer Vision
  (ICCV)}, \bibinfo{year}{2005}, pp. \bibinfo{pages}{1508--1515}.
  \DOIprefix\doi{10.1109/ICCV.2005.104}.
\bibitem[{Rosten and Drummond(2006)}]{Rosten2006}
\bibinfo{author}{E.~Rosten}, \bibinfo{author}{T.~Drummond},
\newblock \bibinfo{title}{Machine learning for high-speed corner detection},
\newblock in: \bibinfo{booktitle}{European Conference on Computer Vision
  (ECCV)}, volume \bibinfo{volume}{3951} of \textit{\bibinfo{series}{Lecture
  Notes in Computer Science}}, \bibinfo{year}{2006}, pp.
  \bibinfo{pages}{430--443}. \DOIprefix\doi{10.1007/11744023_34}.
\bibitem[{Vedaldi and Fulkerson(2008)}]{Vedaldi2008}
\bibinfo{author}{A.~Vedaldi}, \bibinfo{author}{B.~Fulkerson},
  \bibinfo{title}{{VLFeat}: An open and portable library of computer vision
  algorithms}, \bibinfo{year}{2008}. \URLprefix \url{http://www.vlfeat.org}.
\bibitem[{Heinly et~al.(2012)Heinly, Dunn, and Frahm}]{Heinly2012}
\bibinfo{author}{J.~Heinly}, \bibinfo{author}{E.~Dunn}, \bibinfo{author}{J.-M.
  Frahm},
\newblock \bibinfo{title}{Comparative evaluation of binary features},
\newblock in: \bibinfo{booktitle}{European Conference on Computer Vision
  (ECCV)}, volume \bibinfo{volume}{7573} of \textit{\bibinfo{series}{Lecture
  Notes in Computer Science}}, \bibinfo{year}{2012}, pp.
  \bibinfo{pages}{759--773}. \DOIprefix\doi{10.1007/978-3-642-33709-3_54}.
\bibitem[{Itseez(2017)}]{Itseez2017}
\bibinfo{author}{Itseez}, \bibinfo{title}{Open source computer vision library
  ({OpenCV})}, \bibinfo{year}{2017}. \URLprefix
  \url{https://github.com/itseez/opencv}.
\bibitem[{J{\'{e}}gou et~al.(2012)J{\'{e}}gou, Perronnin, Douze, S{\'{a}}nchez,
  P{\'{e}}rez, and Schmid}]{Jegou12ALI}
\bibinfo{author}{H.~J{\'{e}}gou}, \bibinfo{author}{F.~Perronnin},
  \bibinfo{author}{M.~Douze}, \bibinfo{author}{J.~S{\'{a}}nchez},
  \bibinfo{author}{P.~P{\'{e}}rez}, \bibinfo{author}{C.~Schmid},
\newblock \bibinfo{title}{Aggregating local image descriptors into compact
  codes},
\newblock \bibinfo{journal}{{IEEE} Transactions on Pattern Analysis and Machine
  Intelligence} \bibinfo{volume}{34} (\bibinfo{year}{2012})
  \bibinfo{pages}{1704--1716}. \DOIprefix\doi{10.1109/TPAMI.2011.235}.
\bibitem[{Peng et~al.(2015)Peng, Wang, Wang, and Qiao}]{Peng15}
\bibinfo{author}{X.~Peng}, \bibinfo{author}{L.~Wang},
  \bibinfo{author}{X.~Wang}, \bibinfo{author}{Y.~Qiao},
\newblock \bibinfo{title}{Bag of visual words and fusion methods for action
  recognition: Comprehensive study and good practice},
\newblock \bibinfo{journal}{Computer Vision and Image Understanding}
  \bibinfo{volume}{150} (\bibinfo{year}{2015}) \bibinfo{pages}{109--125}.
  \DOIprefix\doi{10.1016/j.cviu.2016.03.013}.
\bibitem[{Gong et~al.(2014)Gong, Wang, Guo, and Lazebnik}]{Gong14MSO}
\bibinfo{author}{Y.~Gong}, \bibinfo{author}{L.~Wang}, \bibinfo{author}{R.~Guo},
  \bibinfo{author}{S.~Lazebnik},
\newblock \bibinfo{title}{Multi-scale orderless pooling of deep convolutional
  activation features},
\newblock in: \bibinfo{booktitle}{European Conference on Computer Vision
  (ECCV)}, volume \bibinfo{volume}{8695}, \bibinfo{year}{2014}, pp.
  \bibinfo{pages}{392--407}. \DOIprefix\doi{10.1007/978-3-319-10584-0_26}.
\bibitem[{Ng et~al.(2015)Ng, Yang, and Davis}]{Ng15}
\bibinfo{author}{J.~Y.~H. Ng}, \bibinfo{author}{F.~Yang},
  \bibinfo{author}{L.~S. Davis},
\newblock \bibinfo{title}{Exploiting local features from deep networks for
  image retrieval},
\newblock in: \bibinfo{booktitle}{Conference on Computer Vision and Pattern
  Recognition Workshops (CVPRW)}, \bibinfo{year}{2015}, pp.
  \bibinfo{pages}{53--61}. \DOIprefix\doi{10.1109/CVPRW.2015.7301272}.
\bibitem[{Paulin et~al.(2016)Paulin, Mairal, Douze, Harchaoui, Perronnin, and
  Schmid}]{Paulin16}
\bibinfo{author}{M.~Paulin}, \bibinfo{author}{J.~Mairal},
  \bibinfo{author}{M.~Douze}, \bibinfo{author}{Z.~Harchaoui},
  \bibinfo{author}{F.~Perronnin}, \bibinfo{author}{C.~Schmid},
\newblock \bibinfo{title}{Convolutional patch representations for image
  retrieval: An unsupervised approach},
\newblock \bibinfo{journal}{International Journal of Computer Vision}
  \bibinfo{volume}{121} (\bibinfo{year}{2016}) \bibinfo{pages}{149--168}.
  \DOIprefix\doi{10.1007/s11263-016-0924-3}.
\bibitem[{Christlein et~al.(2017)Christlein, Gropp, Fiel, and
  Maier}]{Christlein2017}
\bibinfo{author}{V.~Christlein}, \bibinfo{author}{M.~Gropp},
  \bibinfo{author}{S.~Fiel}, \bibinfo{author}{A.~K. Maier},
\newblock \bibinfo{title}{Unsupervised feature learning for writer
  identification and writer retrieval},
\newblock in: \bibinfo{booktitle}{International Conference on Document Analysis
  and Recognition (ICDAR)}, volume~\bibinfo{volume}{1}, \bibinfo{year}{2017},
  pp. \bibinfo{pages}{991--997}. \DOIprefix\doi{10.1109/ICDAR.2017.165}.
\bibitem[{Sculley(2010)}]{Sculley10}
\bibinfo{author}{D.~Sculley},
\newblock \bibinfo{title}{Web-scale \(k\)-means clustering},
\newblock in: \bibinfo{booktitle}{International Conference on World Wide Web
  (WWW)}, \bibinfo{year}{2010}, pp. \bibinfo{pages}{1177--1178}.
  \DOIprefix\doi{10.1145/1772690.1772862}.
\bibitem[{J{\'{e}}gou and Ond{{\v{r}}}ej(2012)}]{Jegou12NEA}
\bibinfo{author}{H.~J{\'{e}}gou}, \bibinfo{author}{C.~Ond{{\v{r}}}ej},
\newblock \bibinfo{title}{Negative evidences and co-occurences in image
  retrieval: The benefit of {PCA} and whitening},
\newblock in: \bibinfo{booktitle}{European Conference on Computer Vision
  (ECCV)}, volume \bibinfo{volume}{7573} of \textit{\bibinfo{series}{Lecture
  Notes in Computer Science}}, \bibinfo{year}{2012}, pp.
  \bibinfo{pages}{774--787}. \DOIprefix\doi{10.1007/978-3-642-33709-3_55}.
\bibitem[{Kessy et~al.(2016)Kessy, Lewin, and Strimmer}]{Kessy2018}
\bibinfo{author}{A.~Kessy}, \bibinfo{author}{A.~Lewin},
  \bibinfo{author}{K.~Strimmer}, \bibinfo{title}{Optimal Whitening and
  Decorrelation}, \bibinfo{type}{e-print}, \bibinfo{year}{2016}.
  \href{http://arxiv.org/abs/1512.00809}{{\tt arXiv:1512.00809}}.
\bibitem[{Fan et~al.(2008)Fan, Chang, Hsieh, Wang, and Lin}]{Fan2008}
\bibinfo{author}{R.-E. Fan}, \bibinfo{author}{K.-W. Chang},
  \bibinfo{author}{C.-J. Hsieh}, \bibinfo{author}{X.-R. Wang},
  \bibinfo{author}{C.-J. Lin},
\newblock \bibinfo{title}{\textsc{Liblinear}: A library for large linear
  classification},
\newblock \bibinfo{journal}{The Journal of Machine Learning Research}
  \bibinfo{volume}{9} (\bibinfo{year}{2008}) \bibinfo{pages}{1871--1874}.
\bibitem[{Chang and Lin(2011)}]{Chang2011}
\bibinfo{author}{C.-C. Chang}, \bibinfo{author}{C.-J. Lin},
\newblock \bibinfo{title}{{\textsc{LibSVM}}: A library for support vector
  machines},
\newblock \bibinfo{journal}{ACM Transactions on Intelligent Systems and
  Technology} \bibinfo{volume}{2} (\bibinfo{year}{2011})
  \bibinfo{pages}{27:1--27:27}. \bibinfo{note}{Software available at
  \url{http://www.csie.ntu.edu.tw/~cjlin/libsvm}}.
\bibitem[{van Rijsbergen(1979)}]{Rijsbergen1979}
\bibinfo{author}{C.~J. van Rijsbergen}, \bibinfo{title}{Information Retrieval},
  \bibinfo{edition}{2\textsuperscript{nd}} ed.,
  \bibinfo{publisher}{Butterworth-Heinemann}, \bibinfo{year}{1979}.
\bibitem[{Simonyan and Zisserman(2015)}]{Simonyan2015}
\bibinfo{author}{K.~Simonyan}, \bibinfo{author}{A.~Zisserman},
\newblock \bibinfo{title}{Very deep convolutional networks for large-scale
  image recognition},
\newblock in: \bibinfo{booktitle}{International Conference on Learning
  Representations}, \bibinfo{year}{2015}.
\bibitem[{Deng et~al.(2009)Deng, Dong, Socher, Li, Li, and Fei-Fei}]{Deng2009}
\bibinfo{author}{J.~Deng}, \bibinfo{author}{W.~Dong},
  \bibinfo{author}{R.~Socher}, \bibinfo{author}{L.-J. Li},
  \bibinfo{author}{K.~Li}, \bibinfo{author}{L.~Fei-Fei},
\newblock \bibinfo{title}{{ImageNet}: A large-scale hierarchical image
  database},
\newblock in: \bibinfo{booktitle}{Conference on Computer Vision and Pattern
  Recognition (CVPR)}, \bibinfo{year}{2009}, pp. \bibinfo{pages}{248--255}.
  \DOIprefix\doi{10.1109/CVPR.2009.5206848}.
\bibitem[{Lin et~al.(2013)Lin, Chen, and Yan}]{Lin2013}
\bibinfo{author}{M.~Lin}, \bibinfo{author}{Q.~Chen}, \bibinfo{author}{S.~Yan},
  \bibinfo{title}{Network In Network}, \bibinfo{type}{e-print},
  \bibinfo{year}{2013}. \href{http://arxiv.org/abs/1312.4400}{{\tt
  arXiv:1312.4400}}.
\bibitem[{Girshick et~al.(2014)Girshick, Donahue, Darrell, and
  Malik}]{Girshick2014}
\bibinfo{author}{R.~Girshick}, \bibinfo{author}{J.~Donahue},
  \bibinfo{author}{T.~Darrell}, \bibinfo{author}{J.~Malik},
\newblock \bibinfo{title}{Rich feature hierarchies for accurate object
  detection and semantic segmentation},
\newblock in: \bibinfo{booktitle}{Conference on Computer Vision and Pattern
  Recognition (CVPR)}, \bibinfo{year}{2014}, pp. \bibinfo{pages}{580--587}.
  \DOIprefix\doi{10.1109/CVPR.2014.81}.
\bibitem[{Kingma and Ba(2014)}]{Kingma2014}
\bibinfo{author}{D.~P. Kingma}, \bibinfo{author}{J.~Ba}, \bibinfo{title}{Adam:
  A Method for Stochastic Optimization}, \bibinfo{type}{e-print},
  \bibinfo{year}{2014}. \href{http://arxiv.org/abs/1412.6980}{{\tt
  arXiv:1412.6980}}.
\bibitem[{Chollet et~al.(2015)}]{Chollet2015}
\bibinfo{author}{F.~Chollet}, et~al., \bibinfo{title}{Keras},
  \bibinfo{howpublished}{GitHub}, \bibinfo{year}{2015}. \URLprefix
  \url{https://github.com/keras-team/keras}.
\bibitem[{Abadi et~al.(2015)Abadi, Agarwal, Barham, Brevdo, Chen, Citro,
  Corrado, Davis, Dean, Devin, Ghemawat, Goodfellow, Harp, Irving, Isard, Jia,
  Jozefowicz, Kaiser, Kudlur, Levenberg, Man\'{e}, Monga, Moore, Murray, Olah,
  Schuster, Shlens, Steiner, Sutskever, Talwar, Tucker, Vanhoucke, Vasudevan,
  Vi\'{e}gas, Vinyals, Warden, Wattenberg, Wicke, Yu, and Zheng}]{Abadi2015}
\bibinfo{author}{M.~Abadi}, \bibinfo{author}{A.~Agarwal},
  \bibinfo{author}{P.~Barham}, \bibinfo{author}{E.~Brevdo},
  \bibinfo{author}{Z.~Chen}, \bibinfo{author}{C.~Citro}, \bibinfo{author}{G.~S.
  Corrado}, \bibinfo{author}{A.~Davis}, \bibinfo{author}{J.~Dean},
  \bibinfo{author}{M.~Devin}, \bibinfo{author}{S.~Ghemawat},
  \bibinfo{author}{I.~Goodfellow}, \bibinfo{author}{A.~Harp},
  \bibinfo{author}{G.~Irving}, \bibinfo{author}{M.~Isard},
  \bibinfo{author}{Y.~Jia}, \bibinfo{author}{R.~Jozefowicz},
  \bibinfo{author}{L.~Kaiser}, \bibinfo{author}{M.~Kudlur},
  \bibinfo{author}{J.~Levenberg}, \bibinfo{author}{D.~Man\'{e}},
  \bibinfo{author}{R.~Monga}, \bibinfo{author}{S.~Moore},
  \bibinfo{author}{D.~Murray}, \bibinfo{author}{C.~Olah},
  \bibinfo{author}{M.~Schuster}, \bibinfo{author}{J.~Shlens},
  \bibinfo{author}{B.~Steiner}, \bibinfo{author}{I.~Sutskever},
  \bibinfo{author}{K.~Talwar}, \bibinfo{author}{P.~Tucker},
  \bibinfo{author}{V.~Vanhoucke}, \bibinfo{author}{V.~Vasudevan},
  \bibinfo{author}{F.~Vi\'{e}gas}, \bibinfo{author}{O.~Vinyals},
  \bibinfo{author}{P.~Warden}, \bibinfo{author}{M.~Wattenberg},
  \bibinfo{author}{M.~Wicke}, \bibinfo{author}{Y.~Yu},
  \bibinfo{author}{X.~Zheng}, \bibinfo{title}{{TensorFlow}: Large-scale machine
  learning on heterogeneous systems}, \bibinfo{year}{2015}. \URLprefix
  \url{https://www.tensorflow.org}.
\bibitem[{Buerhop-Lutz et~al.(2018)Buerhop-Lutz, Deitsch, Maier, Gallwitz,
  Berger, Doll, Hauch, Camus, and Brabec}]{Buerhop2018}
\bibinfo{author}{C.~Buerhop-Lutz}, \bibinfo{author}{S.~Deitsch},
  \bibinfo{author}{A.~Maier}, \bibinfo{author}{F.~Gallwitz},
  \bibinfo{author}{S.~Berger}, \bibinfo{author}{B.~Doll},
  \bibinfo{author}{J.~Hauch}, \bibinfo{author}{C.~Camus},
  \bibinfo{author}{C.~J. Brabec},
\newblock \bibinfo{title}{A benchmark for visual identification of defective
  solar cells in electroluminescence imagery},
\newblock in: \bibinfo{booktitle}{35\textsuperscript{th} European PV Solar
  Energy Conference and Exhibition}, \bibinfo{year}{2018}, pp.
  \bibinfo{pages}{1287--1289}.
  \DOIprefix\doi{10.4229/35thEUPVSEC20182018-5CV.3.15}.
\bibitem[{King and Zeng(2001)}]{King2001}
\bibinfo{author}{G.~King}, \bibinfo{author}{L.~Zeng},
\newblock \bibinfo{title}{Logistic regression in rare events data},
\newblock \bibinfo{journal}{Political Analysis} \bibinfo{volume}{9}
  (\bibinfo{year}{2001}) \bibinfo{pages}{137--163}.
  \DOIprefix\doi{10.1093/oxfordjournals.pan.a004868}.
\bibitem[{Fawcett(2006)}]{Fawcett2006}
\bibinfo{author}{T.~Fawcett},
\newblock \bibinfo{title}{An introduction to {ROC} analysis},
\newblock \bibinfo{journal}{Pattern Recognition Letters} \bibinfo{volume}{27}
  (\bibinfo{year}{2006}) \bibinfo{pages}{861--874}.
  \DOIprefix\doi{10.1016/j.patrec.2005.10.010}.
\bibitem[{Sun et~al.(2017)Sun, Shrivastava, Singh, and Gupta}]{Sun2017}
\bibinfo{author}{C.~Sun}, \bibinfo{author}{A.~Shrivastava},
  \bibinfo{author}{S.~Singh}, \bibinfo{author}{A.~Gupta},
\newblock \bibinfo{title}{Revisiting unreasonable effectiveness of data in deep
  learning era},
\newblock in: \bibinfo{booktitle}{2017 IEEE International Conference on
  Computer Vision (ICCV)}, \bibinfo{year}{2017}, pp. \bibinfo{pages}{843--852}.
  \DOIprefix\doi{10.1109/ICCV.2017.97}.
\bibitem[{van~der Maaten and Hinton(2008)}]{Maaten2008}
\bibinfo{author}{L.~van~der Maaten}, \bibinfo{author}{G.~Hinton},
\newblock \bibinfo{title}{Visualizing data using \(t\)-{SNE}},
\newblock \bibinfo{journal}{The Journal of Machine Learning Research}
  \bibinfo{volume}{9} (\bibinfo{year}{2008}) \bibinfo{pages}{2579--2605}.
\bibitem[{van~der Maaten(2014)}]{Maaten2014}
\bibinfo{author}{L.~van~der Maaten},
\newblock \bibinfo{title}{Accelerating \(t\)-{SNE} using tree-based
  algorithms},
\newblock \bibinfo{journal}{The Journal of Machine Learning Research}
  \bibinfo{volume}{15} (\bibinfo{year}{2014}) \bibinfo{pages}{3221--3245}.
\bibitem[{Zhou et~al.(2016)Zhou, Khosla, Lapedriza, Oliva, and
  Torralba}]{Zhou2016}
\bibinfo{author}{B.~Zhou}, \bibinfo{author}{A.~Khosla},
  \bibinfo{author}{A.~Lapedriza}, \bibinfo{author}{A.~Oliva},
  \bibinfo{author}{A.~Torralba},
\newblock \bibinfo{title}{Learning deep features for discriminative
  localization},
\newblock in: \bibinfo{booktitle}{Conference on Computer Vision and Pattern
  Recognition (CVPR)}, \bibinfo{year}{2016}, pp. \bibinfo{pages}{2921--2929}.
  \DOIprefix\doi{10.1109/CVPR.2016.319}.
\bibitem[{Selvaraju et~al.(2017)Selvaraju, Cogswell, Das, Vedantam, Parikh, and
  Batra}]{Selvaraju2017}
\bibinfo{author}{R.~R. Selvaraju}, \bibinfo{author}{M.~Cogswell},
  \bibinfo{author}{A.~Das}, \bibinfo{author}{R.~Vedantam},
  \bibinfo{author}{D.~Parikh}, \bibinfo{author}{D.~Batra},
\newblock \bibinfo{title}{{Grad-CAM}: Visual explanations from deep networks
  via gradient-based localization},
\newblock in: \bibinfo{booktitle}{IEEE International Conference on Computer
  Vision (ICCV)}, \bibinfo{year}{2017}, pp. \bibinfo{pages}{618--626}.
  \DOIprefix\doi{10.1109/ICCV.2017.74}.
\bibitem[{Chattopadhay et~al.(2018)Chattopadhay, Sarkar, Howlader, and
  Balasubramanian}]{Chattopadhay2018}
\bibinfo{author}{A.~Chattopadhay}, \bibinfo{author}{A.~Sarkar},
  \bibinfo{author}{P.~Howlader}, \bibinfo{author}{V.~N. Balasubramanian},
\newblock \bibinfo{title}{{Grad-CAM++}: Generalized gradient-based visual
  explanations for deep convolutional networks},
\newblock in: \bibinfo{booktitle}{Winter Conference on Applications of Computer
  Vision (WACV)}, \bibinfo{year}{2018}, pp. \bibinfo{pages}{839--847}.
  \DOIprefix\doi{10.1109/WACV.2018.00097}.
\bibitem[{Goodfellow et~al.(2016)Goodfellow, Bengio, and
  Courville}]{Goodfellow2016}
\bibinfo{author}{I.~Goodfellow}, \bibinfo{author}{Y.~Bengio},
  \bibinfo{author}{A.~Courville}, \bibinfo{title}{Deep Learning},
  \bibinfo{publisher}{MIT Press}, \bibinfo{year}{2016}.

\end{thebibliography}

\end{document}